%% file: notes.tex
\documentclass[11pt]{article}

\usepackage[utf8]{inputenc}
\pdfoutput=1
\usepackage[margin=1.1in,letterpaper]{geometry}
\usepackage[round]{natbib}

\usepackage{booktabs}            
\usepackage{multirow}            
\usepackage{amsfonts}            
\usepackage{graphicx}            
\usepackage{duckuments}          
\usepackage{algorithm}
\usepackage{algpseudocode}
\usepackage{etoc}

\usepackage{wrapfig}

\usepackage{utfsym}

\makeatletter
\newlength\aftertitskip     \newlength\beforetitskip
\newlength\interauthorskip  \newlength\aftermaketitskip

\def\maketitle{\par
 \begingroup
   \def\thefootnote{\fnsymbol{footnote}}
   \def\@makefnmark{\hbox to 4pt{$^{\@thefnmark}$\hss}}
   \@maketitle \@thanks
 \endgroup
\setcounter{footnote}{0}
 \let\maketitle\relax \let\@maketitle\relax
 \gdef\@thanks{}\gdef\@author{}\gdef\@title{}\let\thanks\relax}

\def\@startauthor{\noindent \normalsize\bf}
\def\@endauthor{}
\def\@starteditor{\noindent \small {\bf Editor:~}}
\def\@endeditor{\normalsize}
\def\@maketitle{\vbox{\hsize\textwidth
 \linewidth\hsize \vskip \beforetitskip
 {\begin{center} \LARGE\@title \par \end{center}} \vskip \aftertitskip
 {\def\and{\unskip\enspace{\rm and}\enspace}%
  \def\addr{\small\it}%
  \def\email{\hfill\small\tt}%
  \def\name{\normalsize\bf}%
  \def\AND{\@endauthor\rm\hss \vskip \interauthorskip \@startauthor}
  \@startauthor \@author \@endauthor}
}}

\makeatother

\pdfoutput=1                    

\usepackage{amsmath,amsthm}
\usepackage{graphicx}
\usepackage{color}
\usepackage{amssymb,amsfonts,amsxtra,mathrsfs,bm}
\usepackage{multicol}
\usepackage{multirow}
\usepackage{paralist}
\usepackage{xspace}
\usepackage{algorithm}
\usepackage{algpseudocode}
\usepackage[colorlinks=true, urlcolor = blue]{hyperref}
\usepackage{bbm}
\usepackage{nicefrac}
\input{macros.tex}

\title{Mitigating Over-Smoothing and Over-Squashing using Augmentations of Forman-Ricci Curvature}

\author{\name Lukas Fesser
\email{lukas\_fesser@fas.harvard.edu}\\
  \addr{Harvard University}\\
  \name Melanie Weber \email{mweber@seas.harvard.edu}\\
  \addr{Harvard University}
}

\begin{document}
\maketitle

\begin{abstract}
\noindent While Graph Neural Networks (GNNs) have been successfully leveraged for learning on graph-structured data across domains, several potential pitfalls have been described recently. Those include the inability to accurately leverage information encoded in long-range connections (\emph{over-squashing}), as well as difficulties distinguishing the learned representations of nearby nodes with growing network depth (\emph{over-smoothing}). An effective way to characterize both effects is discrete curvature: Long-range connections that underlie over-squashing effects have low curvature, whereas edges that contribute to over-smoothing have high curvature. This observation has given rise to \emph{rewiring} techniques, which add or remove edges to mitigate over-smoothing and over-squashing. Several rewiring approaches utilizing graph characteristics, such as curvature or the spectrum of the graph Laplacian, have been proposed. However, existing methods, especially those based on curvature, often require expensive subroutines and careful hyperparameter tuning, which limits their applicability to large-scale graphs. Here we propose a rewiring technique based on Augmented Forman-Ricci curvature (AFRC), a scalable curvature notation, which can be computed in linear time. We prove that AFRC effectively characterizes over-smoothing and over-squashing effects in message-passing GNNs. We complement our theoretical results with experiments, which demonstrate that the proposed approach achieves state-of-the-art performance while significantly reducing the computational cost in comparison with other methods. Utilizing fundamental properties of discrete curvature, we propose effective heuristics for hyperparameters in curvature-based rewiring, which avoids expensive hyperparameter searches, further improving the scalability of the proposed approach. \footnote{Code available at \url{https://github.com/Weber-GeoML/AFRC_Rewiring}}
\end{abstract}


\section{Introduction}
Graph-structured data is ubiquitous in data science and machine learning applications across domains.  Message-passing Graph Neural Networks (GNNs) have emerged as a powerful architecture for Deep Learning on graph-structured data, leading to many recent success stories in a wide range of disciplines, including biochemistry~\citep{gligorijevic2021structure}, drug discovery~\citep{zitnik}, recommender systems~\citep{wu2022graph} and particle physics~\citep{shlomi2020graph}. 
However, recent literature has uncovered limitations in the representation power of GNNs~\citep{xu2018powerful}, many of which stem from or are amplified by the inability of message-passing graph neural networks to accurately leverage information encoded in long-range connections (\emph{over-squashing}~\citep{alon2021on}), as well as difficulties distinguishing the learned representations of nearby nodes with growing network depth (\emph{over-smoothing}~\citep{li2018deeper}).  As a result, there has been a surge of interest in characterizing over-squashing and over-smoothing mathematically and in developing tools for mitigating both effects.  \\

\noindent Among the two,  over-squashing has received the widest attention,  driven by the importance of leveraging information encoded in long-range connections in both node- and graph-level tasks.   Over-smoothing has been observed to impact, in particular,  the performance of GNNs on node-level tasks.  Characterizations of over-squashing and over-smoothing frequently utilize tools from Discrete Geometry, such as the spectrum of the Graph Laplacian~\citep{banerjee2022oversquashing,karhadkar2022fosr,black2023understanding} or discrete Ricci curvature~\citep{topping_understanding_2022,nguyen2023revisiting}.  Discrete curvature has been linked previously to graph topology and "information flow" in graphs, which has given rise to several applications in network analysis and machine learning~\citep{ni2019community,sia2019ollivier,weber2017characterizing,fesser,tian}.  Classical Ricci curvature characterizes local volume growth rates on manifolds (\emph{geodesic dispersion}); analogous notions in discrete spaces were introduced by Ollivier~\citep{Ol2}, Forman~\citep{forman} and Maas and Erbar~\citep{erbar}, among others.  \\

\noindent Previous work on characterizing over-squashing and over-smoothing with discrete Ricci curvature has utilized Ollivier's notion~\citep{nguyen2023revisiting} (short ORC), as well as a variation of Forman's notion~\citep{topping_understanding_2022} (FRC).  Utilizing ORC is a natural approach, thanks to fundamental relations to the spectrum of the Graph Laplacian and other fundamental graph characteristics~\citep{jost-liu}, which are known to characterize information flow on graphs.  However,  computing ORC requires solving an optimal transport problem for each edge in the graph,  making a characterization of over-squashing and over-smoothing effects via ORC prohibitively expensive on large-scale graphs.  In contrast, FRC can be computed efficiently even on massive graphs due to its simple combinatorial form.  Recent work on \emph{augmentations} of Forman's curvature, which incorporate higher-order structural information (e.g., cycles) into the curvature computation,  has demonstrated their efficiency in graph-based learning,  coming close to the accuracy reached by ORC-based methods at a fraction of the computational cost~\citep{fesser,tian}.  In this work, we will demonstrate that augmented Forman curvature (AFRC) allows for an efficient characterization of over-squashing and over-smoothing.  We relate both effects to AFRC theoretically, utilizing novel bounds on AFRC.  \\

\noindent Besides characterizing over-squashing and over-smoothing,  much recent interest has been dedicated to mitigating both effects in GNNs.  \emph{Graph rewiring},  which adds and removes edges to improve the information flow through the network,  has emerged as a promising tool for improving the quality of the learned node embeddings and their utility in downstream tasks.  Building on characterizations of over-squashing (and, to some extend, over-smoothing) discussed above, several rewiring techniques have been proposed~\citep{karhadkar2022fosr,banerjee2022oversquashing,topping_understanding_2022,black2023understanding,nguyen2023revisiting}.  Rewiring is integrated into GNN training as a preprocessing step,  which alters the graph topology before learning node embeddings.  It has been shown to improve accuracy on node- and graph-level tasks, specifically in graphs with long-range connections.  Here, we argue that effective rewiring should only add a small computational overhead to merit its integration into GNN training protocols.  Hence, the computational complexity of the corresponding preprocessing step,  in tandem with the potential need for and cost of hyperparameter searches,  is an important consideration.  We introduce a graph rewiring technique based on augmentations of Forman's curvature (AFR-$k$), which can be computed in linear time and which comes with an efficiently computable heuristic for automating hyperparameter choices,  avoiding the need for costly hyperparameter searches.  Through computational experiments, we demonstrate that AFR-$k$ has comparable or superior performance compared to state-of-the-art approaches, while being more computationally efficient.

\begin{table}[t]
\centering
\small
\begin{tabular}{lcccc}
\hline
Approach & O-Smo & O-Squ & Complexity & Hyperparams \\
\hline
SDRF~\citep{topping_understanding_2022} & $\usym{2717}$ & $\usym{2713}$ & $O(m \; d_{\max}^2)$  & grid-search   \\
RLEF~\citep{banerjee2022oversquashing} & $\usym{2717}$ & $\usym{2713}$  & $O\big(n^2 \sqrt{\log(n)} \; d_{\max} \big)$  & grid-search   \\
FoSR~\citep{karhadkar2022fosr} & $\usym{2717}$ & $\usym{2713}$ & $O(n^2)$  & grid-search   \\
BORF~\citep{nguyen2023revisiting} & $\usym{2713}$ & $\usym{2713}$ & $O(m \; d_{\max}^3)$ & grid-search \\
GTR~\citep{black2023understanding} & $\usym{2717}$ & $\usym{2713}$ & $O\big(m \; {\rm poly}\log n + n^2 {\rm poly} \log n \big)$ & grid-search  \\
\textbf{AFR-3 (this paper)} & $\usym{2713}$ & $\usym{2713}$ & $O(m \; d_{\max})$ & \textbf{heuristic}  \\
\hline
\end{tabular}
\caption{Comparison of state-of-the art rewiring approaches for mitigating over-squashing (O-Squ) and over-smoothing (O-Smo) regarding complexity and choice of hyperparameters ($n$ denoting the number of vertices, $m$ the number of edges, $d_{\max}$ the maximal node degree).}
\label{tab:rewiring}
\end{table}
\paragraph{Related Work.}
A growing body of literature considers the representational power of GNNs~\citep{xu2018powerful,morris2019weisfeiler} and related structural effects~\citep{alon2021on,li2018deeper,di-giovanni23a,cai2020note,Oono2020Graph,rusch2022graph}.  This includes in particular challenges in leveraging information encoded in long-range connections (\emph{over-squashing}~\citep{alon2021on}), as well as difficulties distinguishing representations of nearby nodes with growing network depth (\emph{over-smoothing}~\citep{li2018deeper}).  
Rewiring has emerged as a popular tool for mitigating over-squashing and over-smoothing.  Methods based on the spectrum of the graph Laplacian~\citep{karhadkar2022fosr}, effective resistance~\citep{black2023understanding}, expander graphs~\citep{banerjee2022oversquashing,deac22a} and discrete curvature~\citep{topping_understanding_2022,nguyen2023revisiting} have been proposed (see Tab.~\ref{tab:rewiring}).  A connection between over-squashing and discrete curvature was first established in~\citep{topping_understanding_2022}. To the best of our knowledge,  augmented Forman curvature has not been considered as a means for studying and mitigating over-squashing and over-smoothing in previous literature. Notably, the \emph{balanced Forman curvature} studied in~\citep{topping_understanding_2022} is distinct from the notion considered in this paper.  Forman's curvature, as well as discrete curvature more generally,  has previously been used in graph-based machine learning, including for unsupervised node clustering (community detection)~\citep{ni2019community,sia2019ollivier,tian,fesser}, graph coarsening~\citep{weber2017curvature} and in representation learning~\citep{lubold_identifying_2022,weber}.  

\paragraph{Contributions.}
Our main contributions are as follows:
\begin{itemize}
    \item We prove that augmentations of Forman's Ricci curvature (AFRC) allow for an effective characterization of over-smoothing and over-squashing effects in message-passing GNNs. 
	 \item We propose an AFRC-based graph rewiring approach (AFR-$k$, Alg.~\ref{alg:cap}),  which is both scalable and performs competitively compared to state-of-the-art rewiring approaches on node- and graph-level tasks. Importantly, AFR-3  can be computed in linear time, allowing for application of the approach to large-scale graphs.
    \item We introduce a novel heuristic for choosing how many edges to add (or remove) in curvature-based rewiring techniques grounded in community detection. This heuristic allows for an effective implementation of our proposed approach, avoiding costly hyperparameter searches.  Notably, the heuristic applies also to existing curvature-based rewiring approaches and hence may be of independent interest.
\end{itemize}

\section{Background and Notation}
\subsection{Message-Passing Graph Neural Networks}
\label{subsection:MPNNs}

Many popular architectures for Graph Machine Learning utilize the \emph{Message-Passing paradigm}~\citep{gori2005new,Hamilton:2017tp}.  Message-passing Graph Neural Networks (short \emph{GNNs}) iteratively compute node representations as a function of the representations of their neighbors with node attributes in the input graph determining the node representation at initialization.  We can view each iteration as a \emph{GNN layer}; the number of iterations performed to compute the final representation can be thought of as the \emph{depth} of the GNN.  Formally, if we let $\mathbf{X}_u^k$ denote the features of node $u$ at layer $k$, then a general formulation for a message-passing Graph Neural Network is 
\begin{equation*}
\mathbf{X}_u^{k+1} = \phi_k \left( \bigoplus_{p \in \tilde{\mathcal{N}}_u} \psi_k \left ( \mathbf{X}_p^k\right)\right)    
\end{equation*}

\noindent Here, $\phi_k$ denotes an update function, $\bigoplus$ an aggregation function, and $\psi_k$ a message function. $\tilde{\mathcal{N}}_u = \mathcal{N}_u \cup \{u\}$ is the extended neighborhood of $u$. Widely used examples of this general formulation include GCN~\citep{Kipf:2017tc}, GIN~\citep{xu2018powerful} and GAT~\citep{Velickovic:2018we}.


\subsection{Discrete Ricci Curvature on Graphs}
\begin{wrapfigure}{R}{0.5\linewidth}
\centering
    \includegraphics[scale=0.4]{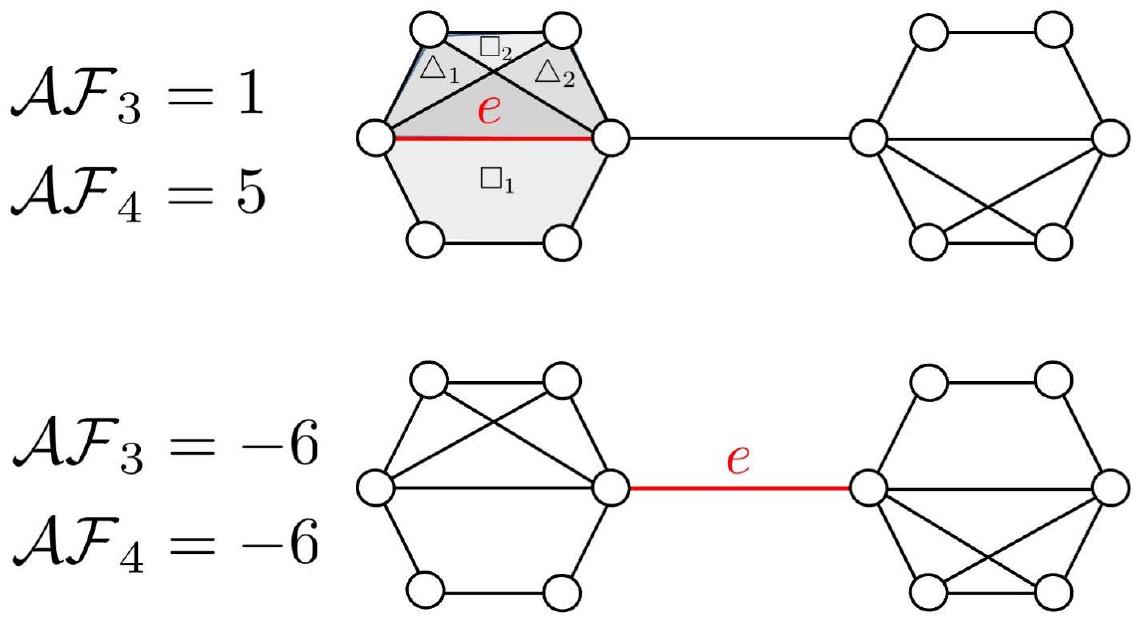}
    \caption{Augmented Forman-Ricci Curvature.}
    \label{fig:afrc}
\end{wrapfigure}
Ricci curvature is a classical tool in Differential Geometry, which establishes a connection between the geometry of the manifold and local volume growth.  Discrete notions of curvature have been proposed via \emph{curvature analogies}, i.e.,  notions that maintain classical relationships with other geometric characteristics.  Forman~\citep{forman} introduced a notion of curvature on CW complexes, which allows for a discretization of a crucial relationship between Ricci curvature and Laplacians, the Bochner-Weizenb{\"o}ck equation.  Here, we utilize an edge-level version of Forman's curvature, which also allows for evaluating curvature contributions of higher-order structures.
Specifically, we consider notions that evaluate higher-order information encoded in cycles of order $\leq k$ (denoted as $\mathcal{AF}_k$), focusing on the cases $k=3$ and $k=4$:
\begin{align}
    \mathcal{AF}_3 (u,v) &= 4 - \deg(u) - \deg(v) + 3 \triangle(u,v)\\
    \mathcal{AF}_4 (u,v) &= 4 - \deg(u) - \deg(v) + 3 \triangle(u,v) + 2 \square(u,v) \; ,
\end{align}
where $\triangle(u,v)$ and $\square(u,v)$ denote the number of triangles and quadrangles containing the edge $(u,v)$ (see Fig.~\ref{fig:afrc} for an example). The derivation of those notions follows directly from~\citep{forman} and can be found, e.g., in~\citep{tian}.

\subsection{Over-squashing and Over-smoothing}

\paragraph{Oversquashing.} It has been observed that bottlenecks in the information flow between distant nodes form as the number of layers in a GNN increases~\citep{alon2021on}.  The resulting information loss can significantly decrease the effectiveness of message-passing and reduce the utility of the learned node representations in node- and graph-level tasks.  This effect is particularly pronounced in long-range connections between distant nodes, such as edges that connect distinct clusters in the graph.  Such edges are characterized by low (negative) Ricci curvature in the sense of Ollivier (ORC),  giving rise to the curvature-based analysis of over-squashing~\citep{topping_understanding_2022,nguyen2023revisiting}. We will show below, that this holds also for AFRC.

\paragraph{Oversmoothing.}
At the same time,  increasing the number of layers can introduce ``shortcuts'' between communities,  induced by the representations of dissimilar nodes (e.g., nodes belonging to different clusters) becoming indistinguishable under the information flow induced by message-passing.  First described by~\citep{li2018deeper}, this effect is known to impact node-level tasks, such as node clustering.  It has been previously shown that oversmoothing arises in positively curved regions of the graph in the sense of Ollivier~\citep{nguyen2023revisiting}.  We will show below, that this holds also for AFRC.

\paragraph{Rewiring.} Graph rewiring describes tools that alter the graph topology by adding or removing edges to improve the information flow in the network.  A taxonomy of state-of-the art rewiring techniques and the AFRC-based approach proposed in this paper with respect to the complexity of the resulting preprocessing step and the choice of hyperparameters can be found in Table~\ref{tab:rewiring}.

\section{Characterizing Over-smoothing and Over-squashing with $\mathcal{AF}_4$}

\noindent Before we can relate the augmented Forman-Ricci curvature of an edge to over-smoothing and over-squashing in GNNs, we require upper and lower bounds on $\mathcal{AF}_3(u, v)$ and $\mathcal{AF}_4(u, v)$. Unlike the Ollivier-Ricci curvature, the augmentations of the Forman-Ricci curvature considered in this paper are not generally bounded independently of the underlying graph. We can prove the following for any (undirected and unweighted) graph $G = (V, E)$:\\

\noindent \textbf{Theorem 3.1.} For $(u, v) \in E$, let $m = \text{deg}(u) \geq \text{deg}(v) = n$. Then for $\mathcal{AF}_3$ and $\mathcal{AF}_4$, we have
\begin{equation}
\begin{split}
4 - m - n &\leq \mathcal{AF}_3(u, v) \leq n + 1 \\
4 - m - n &\leq \mathcal{AF}_4(u, v) \leq 2mn - 3n + 3
\end{split}
\end{equation}

\noindent Note that these edge-specific bounds can easily be extended to graph-level bounds by noting that $n \leq m \leq \left| V\right|-1$. All other results in this section can be extended to graph-level bounds in a similar fashion. The proof for this result can be found in appendix \ref{subsubsection:Thm_3.1}. Our theoretical results and their derivations are largely inspired by \cite{nguyen2023revisiting}, although we require several adjustments such as the bounds above, since we are using the AFRC, not the ORC.

\noindent From the definition in the last section, it is clear that $\mathcal{AF}_4(u, v)$ characterizes how well connected the neighborhoods of $u$ and $v$ are. If many of the nodes in $\mathcal{N}_u$ are connected to $v$ or nodes in $\mathcal{N}_v$, or vice-versa, then $\mathcal{AF}_4$ will be close to its upper bound. Conversely, if $\tilde{\mathcal{N}}_u$ and $\tilde{\mathcal{N}}_v$ have only minimal connections, then $\mathcal{AF}_4(u, v)$ will be close to its lower bound. In the second case, the existing connections will act as bottlenecks and will hinder the message-passing mechanism.

\subsection{$\mathcal{AF}_4$ and Over-smoothing}

\noindent Going back to the definition of the message-passing paradigm in the previous section, we note that at the $k$-th layer of a GNN, every node $p$ sends the same message $\psi_k \left(\mathbf{X}_p^k\right)$ to each node $u$ in its neighborhood. The GNN then aggregates these messages to update the features of node $u$. If $\mathcal{AF}_3$ or $\mathcal{AF}_4$ is very positive, then the neighborhoods of $u$ and $v$ are largely identical, so they receive nearly the same messages, and the difference between their features decreases. The following result makes this intuition for why over-smoothing happens more rigorous. For the proof, see appendix \ref{subsubsection:Thm_3.2}.\\

\noindent \textbf{Theorem 3.2.} Consider an updating rule as defined in section \ref{subsection:MPNNs} and suppose that $\mathcal{AF}_4(u, v) > 0$. For some $k$, assume the update function $\phi_k$ is $L$-Lipschitz, $|\mathbf{X}_p^k| \leq C$ is bounded for all $p \in \mathcal{N}(u) \cup \mathcal{N}(v)$, and the message passing function is bounded, i.e. $|\psi_k(\mathbf{x})| \leq M |\mathbf{x}|, \forall \mathbf{x}$. Let $\bigoplus$ be the sum operation. Then there exists a constant $H > 0$, depending on $L, M$, and $C$, such that 
\begin{equation}
    \left| \mathbf{X}_u^{k + 1} - \mathbf{X}_v^{k + 1}\right| \leq H(2mn - 3n + 3 - \mathcal{AF}_4(u, v))
\end{equation}

\noindent where $m = \text{deg}(u) \geq \text{deg}(v) = n > 1$.\\

\noindent We can prove analogous results for $\mathcal{AF}_3(u, v)$ when $\bigoplus$ is the mean or the sum operation (\ref{subsubsection:Thm_3.2_ext}). Note that Theorem 3.2 applies to most GNN architectures, with the exception of GAT, since GAT uses a learnable weighted mean operation for aggregation. Similar to previous results on the ORC, Theorem 3.2 shows that edges whose augmented Forman-Ricci curvature is close to their upper bound force local node features to become similar. If $\mathcal{AF}_4(e) \approx 2mn - 3n + 3$ for most edges $e$, then we can expect the node features to quickly converge to each other even if the GNN is relatively shallow. For regular graphs, we can extend our analysis and show that the difference between the features of neighboring nodes $u$ and $v$ decays exponentially fast. If in addition, we assume that the diameter of the graph is bounded, then this is true for any pair of nodes $u, v$.\\

\noindent \textbf{Propositon 3.3.} Consider an updating rule as before. Assume the graph is regular. Suppose there exists a constant $\delta$, such that for all edges $(u, v) \in E$, $\mathcal{AF}_4(u, v) \geq \delta > 0$. For all $k$, assume the update functions $\phi_k$ are $L$-Lipschitz, $\bigoplus$ is the mean operation, $|\mathbf{X}_p^0| \leq C$ is bounded for all $p \in V$, and the message passing functions are bounded linear operators, i.e. $|\psi_k(\mathbf{x})| \leq M |\mathbf{x}|, \forall \mathbf{x}$. Then the following inequality holds for $k \geq 1$ and any neighboring vertices $u \sim v$
\begin{equation}
    \left| \mathbf{X}_u^{k} - \mathbf{X}_v^{k}\right| \leq \frac{1}{3}C \left(\frac{3LM(2mn - 3n + 3 -\delta)}{n+1} \right)^k
\end{equation}

\noindent If in addition, the diameter of $G$ is bounded, i.e. $d = \max_{u \in V} \max_{v \in V} d(u, v) \leq D$, then for any $u, v \in V$, not necessarily connected, we have
\begin{equation}
    \left| \mathbf{X}_u^{k} - \mathbf{X}_v^{k}\right| \leq \frac{1}{3}DC \left(\frac{3LM(2mn - 3n + 3 -\delta)}{n + 1} \right)^k
\end{equation}

\noindent For the proof, see appendix \ref{subsubsection:Prop_3.3}. The above result gives us the aforementioned exponential decay in differences between node features, since for appropriately chosen $C_1, C_2 > 0$, we have
\begin{equation}
    \sum_{(u, v) \in E} \left| \mathbf{X}_u^{k} - \mathbf{X}_v^{k}\right| \leq C_1 e^{-C_2 k}
\end{equation}

\noindent Even if most real-word graphs possess negatively curved edges, we still expect an abundance of edges whose AFRC is close to its upper bound to lead to over-smoothing in GNNs.

\subsection{$\mathcal{AF}_4$ and Over-squashing}

\noindent In this subsection, we relate $\mathcal{AF}_4$ to the occurrence of bottlenecks in a graph, which in turn cause the over-squashing phenomenon. Message-passing between neighborhoods requires connections of the form $(p, q)$ for $p \in \tilde{\mathcal{N}}_u\setminus\{v\}$ and $q \in \tilde{\mathcal{N}}_v\setminus\{u\}$. As the next result shows, the number of these connections can be upper-bounded using $\mathcal{AF}_4$. The proof for this can be found in appendix \ref{subsubsection:Prop_3.4}.\\

\noindent \textbf{Proposition 3.4.} Let $(u, v) \in E$ and let $S \subset E$ be the set of all $(p, q)$ with $p \in \tilde{N}_u \setminus \{ v\}$ and $q \in \tilde{N}_v \setminus \{u\}$. Then
\begin{equation}
    \left| S\right| \leq \frac{\mathcal{AF}_4(u, v) + \text{deg}(u) + \text{deg}(v) - 4}{2}
\end{equation}

\noindent An immediate consequence of this result is that if $\mathcal{AF}_4$ is close to its lower bound $4 - \text{deg}(u) - \text{deg}(v)$, then the number of connections between $\tilde{\mathcal{N}}_u$ and $\tilde{\mathcal{N}}_v$ has to be small. Hence these edges will induce bottlenecks in $G$, which in turns causes over-squashing. 

\section{$\mathcal{AF}_3$- and $\mathcal{AF}_4$-based Rewiring}

\subsection{AFR-3 and AFR-4}

\noindent Following our theoretical results in the last section, an $\mathcal{AF}_3$- or $\mathcal{AF}_4$-based rewiring algorithm should remove edges whose curvature is close to the upper bound to avoid over-smoothing, and add additional edges to the neighborhoods of those edges whose curvature is particularly close to the lower bound to mitigate over-squashing. In line with this intuition, we propose \emph{AFR}-$k$, a novel algorithm for graph rewiring that, like the ORC-based BORF, can address over-smoothing and over-squashing simultaneously, while being significantly cheaper to run.\\

\begin{algorithm}[h!]
\caption{AFR-3}\label{alg:cap}
\begin{algorithmic}
\Require A graph $G(V, E)$, heuristic (Bool), \# edges to add $h$, \# edges to remove $l$
\For{$e \in E$}
\State compute $\mathcal{AF}_3(e)$
\EndFor
\State Sort $e \in E$ by $\mathcal{AF}_3(e)$
\If{heuristic == True}
\State - Compute lower threshold $\Delta_L$. For each $(u, v) \in E$ with $\mathcal{AF}_3(u, v) < \Delta_L$, \State   choose $w \in \mathcal{N}_u \setminus \mathcal{N}_v$ uniformly at random, add $(w, v)$ to $E$.
\State - Compute upper threshold $\Delta_U$ and remove all edges $(u, v)$ with $\mathcal{AF}_3(u, v) > \Delta_U$ 
\Else 
\State - Find $h$ edges with lowest $\mathcal{AF}_3$ values, for each of these edges, 
\State choose $w \in \mathcal{N}_u \setminus \mathcal{N}_v$ uniformly at random, add $(w, v)$ to $E$.
\State - Find $k$ edges with highest $\mathcal{AF}_3$ values and remove them from $E$.
\EndIf \\
\Return a rewired graph $G' = (V, E')$ with new edge set $E'$.

\end{algorithmic}
\end{algorithm}

\noindent Assume that we want to add $h$ edges to $E$ to reduce over-squashing, and remove $l$ edges from $E$ to mitigate over-smoothing. Then for AFR-3 without heuristics for edge addition or removal, we first compute $\mathcal{AF}_3(e)$ for all $e \in E$. Then, we sort the the edges by curvature from lowest to highest. For each of the $h$ edges with lowest curvature values, we choose $w \in \mathcal{N}_u \setminus \mathcal{N}_v$ uniformly at random if $\text{deg}(u) \geq \text{deg}(v)$ and add $(w, v)$ to $E$. If no such $w$ exists, we continue on to the next edge. Finally, we remove the $l$ edges with the highest curvature values from $E$. We return the rewired graph $G' = (V, E')$, where $E'$ is the rewired edge set. AFR-4 follows by analogy, only that we use $\mathcal{AF}_4$ instead of $\mathcal{AF}_3$. While $\mathcal{AF}_3$ is cheaper to run than $\mathcal{AF}_4$, both are more economical than their ORC-based cousin BORF, which is due to the next result.\\

\noindent \textbf{Theorem 4.1.} Computing the ORC scales as $O\left(\left|E\right| d_\text{max}^3 \right)$, while computing $\mathcal{AF}_3$ scales as $O\left(\left|E\right| d_\text{max} \right)$, and computing $\mathcal{AF}_4$ scales as $O\left(\left|E\right| d_\text{max}^2 \right)$.\\

\noindent Here, $d_\text{max}$ is the highest degree of a node in $G$. For the proof, see appendix \ref{subsubsection:Thm_4.1}. Note that we could run AFR-3 (AFR-4) as a batch-based algorithm like BORF. For $N$ batches, we would recompute $\mathcal{AF}_3$ ($\mathcal{AF}_4$) $N$ times, and for each batch we would identify the $k$ edges with highest $\mathcal{AF}_3$ ($\mathcal{AF}_4$) values and the $h$ edges with lowest $\mathcal{AF}_3$ ($\mathcal{AF}_4$) values and proceed with them as before. In practice, we find that this rarely improves performance. Results on this are presented in appendix \ref{subsubsection:multiple_rewiring}.

\subsection{Heuristics for adding and removing Edges}

\noindent A natural question to ask upon seeing the AFR algorithm is how many edges one should add or remove for a given graph $G$. For previous methods such as BORF, answering this question involved costly hyperparameter tuning. In this subsection, we use results from community detection in network science to motivate heuristics to automatically determine $h$ and $k$.\\

\noindent \textbf{Curvature Thresholds.} Suppose we are given a graph $G = (V, E)$ and a partition of the network into communities. As previously mentioned, edges within communities tend to have positive Ollivier-Ricci curvature, while edges between communities tend to have negative curvature values. Theoretical evidence for this observation for graphs with community structure has been given in~\citep{ni2019community,gosztolai_unfolding_2021,tian}.\\

\cite{sia2019ollivier} use this observation to propose a community detection algorithm: remove the most negatively curved edge $e$ from $G$, recompute the ORC for all edges sharing a node with $e$ and repeat these two steps until all negatively curved edges have been removed. The connected components in the resulting graph $G'$ are the communities. For $\mathcal{AF}_3$ or $\mathcal{AF}_4$, a natural threshold to distinguish between inter- and intra-community edges is missing, so \cite{fesser} propose to use a Gaussian mixture model with two modes. They fit two normal distributions $\mathcal{N}_1(\mu_1, \sigma_1)$ and $\mathcal{N}_2(\mu_2, \sigma_2)$ with $\mu_1 > \mu_2$ to the curvature distribution and then determine the curvature threshold
\begin{equation}
    \Delta_L=\frac{\sigma_2}{\sigma_1+\sigma_2}\mu_1 + \frac{\sigma_1}{\sigma_1+\sigma_2}\mu_2
\end{equation}

\begin{wrapfigure}{R}{0.5\linewidth}
\centering
    \includegraphics[scale=0.25]{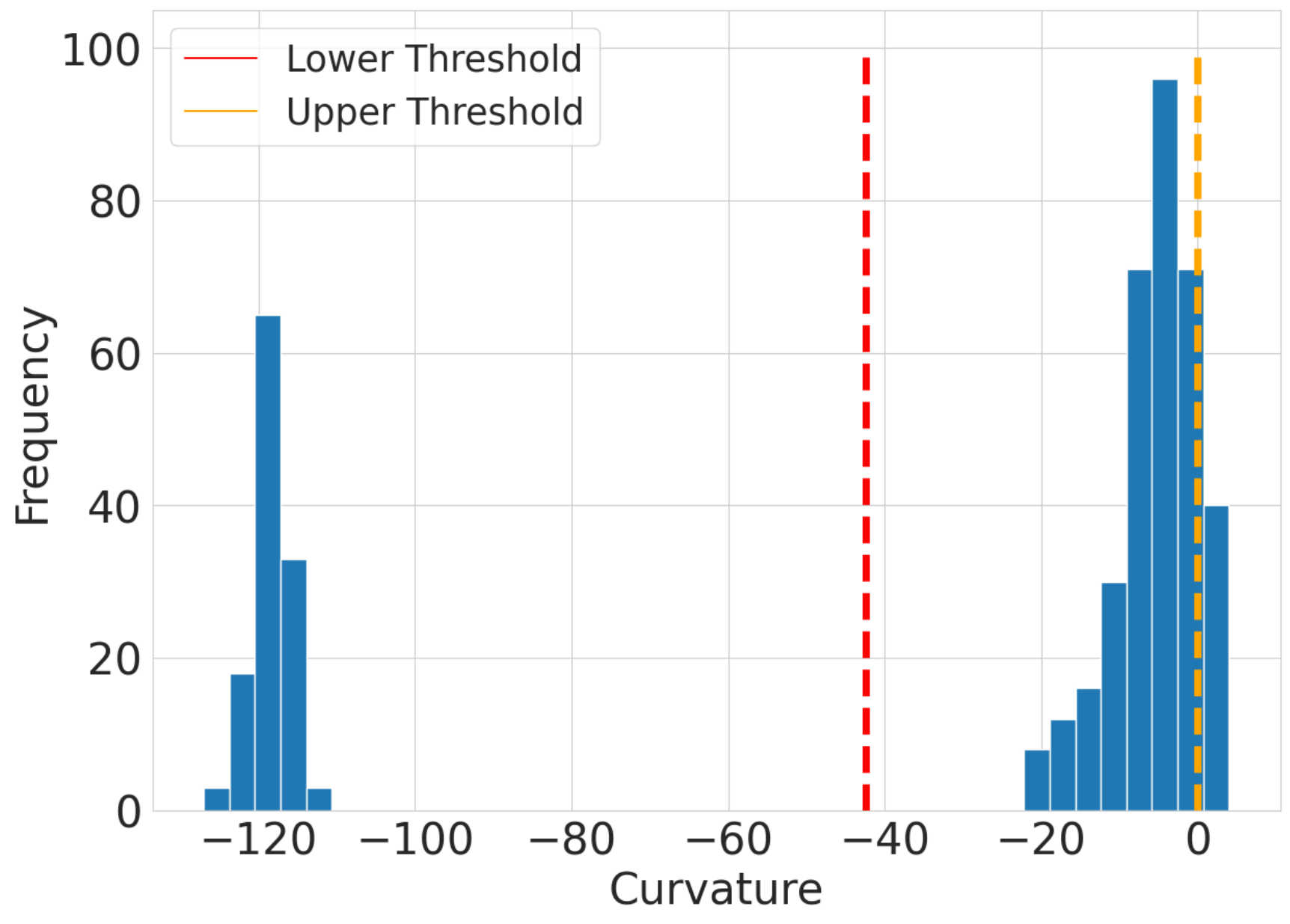}
    \caption{Upper and lower thresholds for the Wisconsin dataset ($\mathcal{AF}_3$).}
    \label{fig:threshold}
\end{wrapfigure}

\noindent Using this as the threshold between inter- and intra-community edges, their $\mathcal{AF}_3$-based sequential community detection algorithm achieves competitive performance while being significantly cheaper to run than the ORC-based original. Based on this, we propose the following heuristic to determine the number $h$ of edges to add for AFR.\\

\noindent \textbf{Heuristic for edge addition.} Instead of choosing the number $h$ of edges to add by hand, we calculate the curvature threshold $\Delta_L$ as above. We then add a new edge for each edge $e$ with $\mathcal{AF}_3(e) < \Delta_L$ ($\mathcal{AF}_4(e) < \Delta_L$) using the same procedure as before.\\

\noindent \textbf{Heuristic for edge removal.} To identify particularly positively curved edges which might lead to over-smoothing, we compute the upper curvature threshold $\Delta_U$ as
\begin{equation}
    \Delta_U = \mu_1 + \sigma_1
\end{equation}

\noindent We remove all edges $e$ with $\mathcal{AF}_3(e) > \Delta_U$ ($\mathcal{AF}_4(e) > \Delta_U)$, i.e. all edges whose curvature is more than one standard deviation above the mean of the normal distribution with higher curvature values.

\section{Experiments}

\noindent In this section, we experimentally demonstrate the effectiveness and computational efficiency of our proposed AFR rewiring algorithm and of our heuristics for edge addition and removal. We first compare AFR against other rewiring alternatives on a variety of tasks, including node and graph classification, as well as long-range tasks. Details on the rewiring strategies we compare against can be found in appendix \ref{subsubsection:other_algos}. We also show that our heuristics allow AFR to achieve competitive performance on all tasks without costly hyperparameter tuning. Our code will be made publicly available upon publication.\\


\noindent \textbf{Experimental details.} Our experiments are designed as follows: for a given rewiring strategy, we first apply it as a preprocessing step to all graphs in the datasets considered. We then train a GNN on a part of the rewired graphs and evaluate its performance on a withheld set of test graphs. We use a train/val/test split of 50/25/25. As GNN architectures, we consider GCN and GIN. Settings and optimization hyper-parameters are held constant across tasks and baseline models for all rewiring methods, so we can rule out hyper-parameter tuning as a source of performance gain. When not using our heuristics, we obtain the settings for the individual rewiring option via hyperparameter tuning. The only hyperparameter choice which we do not optimize using grid search is $\tau$ in SDRF, which we set to $\infty$, in line with \cite{nguyen2023revisiting}. For all rewiring methods and hyperparameter choices, we record the test set accuracy of the settings with the highest validation accuracy. As there is a certain stochasticity involved, especially when training GNNs, we accumulate experimental results across 100 random trials. We report the mean test accuracy, along with the $95\%$ confidence interval. Details on all data sets can be found in appendix \ref{subsection:datasets}.

\subsection{Results using hyperparameter search} 

\noindent Tables \ref{tab:GCN_node_hyper} and \ref{tab:GCN_graph_hyper} present the results of our experiments for node and graph classification with hyperparameter tuning. The exact hyperparameter settings for each dataset can be found in appendix \ref{subsubsection:hyperparam}, where we also present additional results using GIN as an architecture (\ref{subsubsection:GIN_node_graph_heur}). For the experiments with GCNs presented here, AFR-3 and AFR-4 outperform all other rewiring strategies and the no-rewiring baseline on all node classification datasets and on four of our five graph classification datasets. We expect AFR-3, AFR-4, and BORF to attain generally higher accuracies than FoSR and SDRF, because unlike FoSR and SDRF, they can address over-smoothing by removing edges.

\begin{table}[h!]\label{tab:afrc_orc_hyperparam}
\centering
\small
\begin{tabular}{|l|c|c|c|c|c|c|}
\hline
 & 
\multicolumn{6}{|c|}{GCN} \\
\hline
DATA SET & AFR-3 & AFR-4 & BORF & SDRF & FOSR & NONE \\
\hline
CORA & $87.5 \pm 0.5$ & $\mathbf{88.1 \pm 0.5}$ & $87.9 \pm 0.7$ & $86.4 \pm 2.1$ & $86.9 \pm 2.0$ & $86.6 \pm 0.8$ \\
CITESEER & $\mathbf{74.4 \pm 1.0}$ & $73.3 \pm 0.6$ & $73.4 \pm 0.6$ & $72.6 \pm 2.2$ & $73.5 \pm 2.0$ & $71.7 \pm 0.7$ \\
TEXAS & $49.7 \pm 0.5$ & $\mathbf{51.4 \pm 0.5}$ & $48.9 \pm 0.5$ & $43.6 \pm 1.2$ & $46.9 \pm 1.2$ & $44.1 \pm 0.5$ \\
CORNELL & $\mathbf{49.7 \pm 3.4}$ & $48.9 \pm 3.3$ & $48.1 \pm 2.9$ & $43.1 \pm 1.2$ & $43.9 \pm 1.1$ & $46.8 \pm 3.0$\\
WISCON. & $47.3 \pm 2.4$ & $\mathbf{52.2 \pm 2.4}$ & $46.5 \pm 2.6$ & $47.1 \pm 1.0$ & $48.5 \pm 1.0$ & $44.6 \pm 2.9$\\
CHAMEL. & $62.3 \pm 0.9$ & $\mathbf{62.5 \pm 0.9}$ & $61.4 \pm 0.9$ & $59.5 \pm 0.4$ & $59.3 \pm 1.9$ & $59.1 \pm 1.4$\\
\hline
COCO & $\mathbf{10.1 \pm 1.1}$ & $9.8 \pm 1.1$ & $10.1 \pm 1.2$ & $8.5 \pm 1.0$ & $9.3 \pm 1.4$ & $7.8 \pm 0.4$\\
PASCAL & $14.3 \pm 1.5$ & $\mathbf{14.4 \pm 1.4}$ & $14.1 \pm 1.1$ & $11.7 \pm 0.9$ & $13.8 \pm 1.3$ & $10.4 \pm 0.6$\\
\hline
\end{tabular}

\caption{Classification accuracies of GCN with AFR-3, AFR-4, BORF, SDRF, FoSR, or no rewiring strategy using best hyperparameters. Highest accuracies on any given dataset are highlighted in bold. We report F1 scores for the LGRB COCO and PASCAL datasets. 
}
\label{tab:GCN_node_hyper}
\end{table}

\noindent \textbf{Comparing AFR-3 and AFR-4.} In section 3, we used $\mathcal{AF}_4$ for our theoretical results due to it allowing us to better analyze over-smoothing. However, for many real-world datasets, $\mathcal{AF}_3$ and $\mathcal{AF}_4$ are highly correlated \cite{fesser}. We find that this also holds true for the datasets considered here (see appendix \ref{subsection:datasets}), which might explain why AFR-3, AFR-4, and BORF all achieve comparable accuracies, with AFR-3 and AFR-4 often performing marginally better. This also suggests that characterizing over-smoothing and over-squashing using $\mathcal{AF}_3$ or $\mathcal{AF}_4$ is sufficient. The computationally more expensive ORC is not needed. In practice, one would always use AFR-3 due to competitive performance and excellent scalability.

\begin{table}[h!]\label{tab:afrc_orc_hyperparam}
\centering
\small
\begin{tabular}{|l|c|c|c|c|c|c|}
\hline
 & 
\multicolumn{6}{|c|}{GCN}\\
\hline
DATA SET & AFR-3 & AFR-4 & BORF & SDRF & FOSR & NONE \\
\hline
MUTAG & $69.7 \pm 2.1$ & $68.7 \pm 1.9$ & $68.2 \pm 2.4$ & $65.1 \pm 2.2$ & $\mathbf{70.0 \pm 2.2}$ & $62.7 \pm 2.1$ \\
ENZYMES & $25.9 \pm 1.2$ & $\mathbf{26.3 \pm 1.2}$ & $26.0 \pm 1.2$ & $24.3 \pm 1.2$ & $24.9 \pm 1.1$ & $25.4 \pm 1.3$ \\
IMDB & $\mathbf{50.4 \pm 0.9}$ & $49.8 \pm 1.1$ & $48.6 \pm 0.9$ & $48.6 \pm 0.9$ & $48.3 \pm 0.9$ & $48.1 \pm 1.0$\\
PROTEINS & $\mathbf{62.7 \pm 0.8}$ & $61.2 \pm 0.9$ & $61.5 \pm 0.7$ & $59.5 \pm 0.8$ & $59.3 \pm 0.9$ & $59.6 \pm 0.9$\\
\hline
PEPTIDES & $\mathbf{44.7 \pm 2.8}$ & $44.4 \pm 2.8$ & $43.9 \pm 2.6$ & $41.8 \pm 1.5$ & $44.3 \pm 2.2$ & $40.5 \pm 2.1$ \\
\hline
\end{tabular}

\caption{Classification accuracies of GCN with AFR-3, AFR-4, BORF, SDRF, FoSR, or no rewiring strategy using best hyperparameters. Highest accuracies on any given dataset are highlighted in bold.}
\label{tab:GCN_graph_hyper}
\end{table}

\subsection{Results using heuristics for edge addition and removal}

\noindent Using the same datasets as before, we now test our heuristics for replacing hyperparameter tuning. We study the effects of using the thresholds proposed in section 4 on AFR-3, AFR-4, and BORF.  Tables \ref{tab:GCN_GIN_node_heur} and \ref{tab:GCN_GIN_graph_heur} present the accuracies attained by GCN and GIN architectures using our heuristics for finding upper and lower thresholds. Comparing these to the results in Tables \ref{tab:GCN_node_hyper} and \ref{tab:GCN_graph_hyper}, we see that our heuristics outperform hyperparameter tuning on four out of six node classification datasets, and are competitive on the other two. Similarly on the graph classification tasks, where our heuristics achieve superior performance on four out of five tasks.

\begin{table}[h!]
\centering
\small
\begin{tabular}{|l|c|c|c|c|c|c|}
\hline
 & 
\multicolumn{3}{|c|}{GCN} & \multicolumn{3}{|c|}{GIN} \\
\hline
DATA SET & AFR-3 & AFR-4 & BORF & AFR-3 & AFR-4 & BORF\\
\hline
CORA & $87.8 \pm 0.7$ & $87.9 \pm 0.9$ & $87.6 \pm 0.7$ & $77.9 \pm 1.2$ & $78.0 \pm 1.1$ & $78.4 \pm 1.1$ \\
CITESEER & $74.6 \pm 0.7$ & $74.7 \pm 0.7$ & $74.2 \pm 0.8$ & $65.1 \pm 0.7$ & $64.7 \pm 0.6$ & $64.6 \pm 0.6$\\
TEXAS & $52.4 \pm 3.3$ & $48.4 \pm 3.2$ & $50.5 \pm 2.8$ & $68.7 \pm 3.1$ & $63.8 \pm 3.8$ & $62.3 \pm 2.0$\\
CORNELL & $50.5 \pm 4.3$ & $46.2 \pm 2.6$ & $44.6 \pm 1.9$ & $51.9 \pm 4.2$ & $47.3 \pm 2.3$ & $48.4 \pm 2.5$\\
WISCON. & $52.4 \pm 2.6$ & $49.0 \pm 2.0$ & $48.2 \pm 3.0$ & $58.4 \pm 2.9$ & $58.7 \pm 2.3$& $53.5 \pm 2.6$\\
CHAMEL. & $62.2 \pm 1.2$ & $62.1 \pm 1.2$ & $61.1 \pm 1.4$ & $67.1 \pm 2.1$ & $65.9 \pm 2.2$& $66.3 \pm 2.2$\\
\hline
COCO & $10.3 \pm 1.3$ & $9.6 \pm 1.2$ & $10.5 \pm 1.2$ & $13.1 \pm 2.1$ & $13.5 \pm 2.2$ & $13.6 \pm 2.2$  \\
PASCAL & $14.2 \pm 1.5$ & $14.3 \pm 1.5$ & $14.8 \pm 1.1$ & $16.0 \pm 1.7$ & $15.8 \pm 1.7$ & $16.4 \pm 1.8$ \\
\hline
\end{tabular}

\caption{Node classification accuracies of GCN and GIN with AFR-3, AFR-4, and BORF using our heuristics to avoid hyperparameter tuning. }
\label{tab:GCN_GIN_node_heur}
\end{table}

\noindent For BORF, we use zero as the lower threshold to identify bottleneck edges. We conduct additional experiments in appendix \ref{subsubsection:orc_mixtures} which show that the lower thresholds which we get from fitting two normal distributions to the ORC distributions are in fact very close to zero on all datasets considered here. Further ablations that study the effects of our two heuristics individually can also be found in appendices \ref{subsubsection:adding_ablations} and \ref{subsubsection:removing_ablations}.\\

\begin{table}[h!]\label{tab:afrc_orc_heuristic}
\centering
\small
\begin{tabular}{|l|c|c|c|c|c|c|}
\hline
 & 
\multicolumn{3}{|c|}{GCN} & \multicolumn{3}{|c|}{GIN} \\
\hline
DATA SET & AFR-3 & AFR-4 & BORF & AFR-3 & AFR-4 & BORF\\
\hline
MUTAG & $71.4 \pm 2.2$ & $69.9 \pm 2.6$ & $68.5 \pm 1.9$ & $70.9 \pm 2.7$ & $73.4 \pm 2.4$ & $75.4 \pm 2.8$ \\
ENZYMES & $26.1 \pm 1.0$ & $25.5 \pm 1.0$ & $23.3 \pm 1.2$ & $37.6 \pm 1.2$ & $32.1 \pm 1.4$ & $31.9 \pm 1.2$ \\
IMDB & $50.1 \pm 0.9$ & $49.0 \pm 0.9$ & $49.4 \pm 1.0$ & $68.9 \pm 1.1$ & $67.8 \pm 1.2$ & $67.7 \pm 1.5$ \\
PROTEINS & $62.2 \pm 0.8$ & $61.2 \pm 0.9$ & $61.0 \pm 0.9$ & $73.0 \pm 1.5$ & $72.7 \pm 1.3$ & $72.3 \pm 1.2$ \\
\hline
PEPTIDES & $44.8 \pm 2.8$ & $43.6 \pm 2.5$ & $44.3 \pm 2.8$ & $49.2 \pm 1.5$ & $50.7 \pm 1.6$ & $50.1 \pm 1.6$\\
\hline
\end{tabular}

\caption{Graph classification accuracies of GCN and GIN with AFR-3, AFR-4, and BORF using our heuristics to avoid hyperparameter tuning.}
\label{tab:GCN_GIN_graph_heur}
\end{table}

\noindent \textbf{Long-range tasks.} We also evaluated AFR-3 and AFR-4 on the Peptides-func graph classification dataset and on the PascalVOC-SP and COCO-SP node classification datasets, which are part of the long-range tasks introduced by \cite{dwivedi2022LRGB}. As Table \ref{tab:GCN_graph_hyper} shows, AFR-3 and AFR-4 outperform all other rewiring methods on the graph classification task and significantly improve on the no-rewiring baseline. Our heuristics are also clearly still applicable, as they result in comparable performance (Table \ref{tab:GCN_GIN_graph_heur}). Our experiments with long-range node classification yield similar results, as can be seen in Tables \ref{tab:GCN_node_hyper} and \ref{tab:GCN_GIN_node_heur}. Additional experiments with the LRGB datasets using GIN can be found in \ref{subsubsection:GIN_node_graph_heur}.

\section{Discussion}
In this paper we have introduced formal characterizations of over-squashing and over-smoothing effects using augementations of Forman's Ricci curvature,  a simple and scalable notion of discrete curvature.  Based on this characterization, we proposed a scalable graph rewiring approach, which exhibits performance comparable or superior to state-of-the-art rewiring approaches on node- and graph-level tasks.  We further introduce an effective heuristic for hyperparameter choices in curvature-based graph rewiring, which removes the need to perform expensive hyperparameter searches.\\

\noindent There are several avenues for further investigation.  We believe that the complexity of rewiring approaches merits careful consideration and should be evaluated in the context of expected performance gains in applications.  This includes in particular the choice of hyperparameters, which for many state-of-the-art rewiring approaches requires expensive grid searches. While we propose an effective heuristic for curvature-based approaches (both existing and proposed herein), we believe that a broader study on transparent hyperparameter choices across rewiring approaches is merited.  While performance gains in node- and graph-level tasks resulting from rewiring have been established empirically,  a mathematical analysis is still largely lacking and an important direction for future work.  Our proposed hyperparameter heuristic is linked to the topology and global geometric properties of the input graph.  We believe that similar connections could be established for rewiring approaches that do not rely on curvature.  Building on this,  a systematic investigation of the suitability of different rewiring and corresponding hyperparameter choices dependent on graph topology would be a valuable direction for further study. Similarly, differences between the effectiveness of rewiring in homophilous vs. heterophilous graphs strike us as an important direction for future work.

\section{Limitations}
The present paper does not explicitly consider the impact of graph topology on the efficiency of AFR-$k$. Specifically, our proposed heuristic implicitly assumes that the graph has community structure, which is true in many, but not all, applications. Our experiments are restricted to one node- and graph-level task each. While this is in line with experiments presented in  related works, a wider range of tasks would give a more complete picture.

\bibliographystyle{plainnat}
\bibliography{reference} 

\clearpage
\appendix

\section{Appendix}

\localtableofcontents

\subsection{Proofs of theoretical results in the main text}

\subsubsection{Theorem 3.1}
\label{subsubsection:Thm_3.1}

\noindent \textit{Proof:} Let $G = (V, E)$ be an unweighted, undirected graph, let $(u, v) \in E$ and let $t$ and $q$ denote the number of triangles and quadrangles, respectively, containing the edge $(u, v)$. The lower bound in both cases ($\mathcal{AF}_3$ and $\mathcal{AF}_4$) follows straight from the definition and the fact that $t$ and $q$ are non-negative by definition. The lower bounds are attained when there are no triangles (resp. no triangles and quadrangles) containing $(u, v)$ in $E$.\\

\noindent For the upper bound on $\mathcal{AF}_3$, note that $t \leq n - 1$ (and $n \geq 1$ by assumption). Also note that the only way to increase $\mathcal{AF}_3(u, v)$ is to add triangles, which increases $\mathcal{AF}_3(u, v)$ by one (as we also increase the degrees of $u$ and $v$ by one each). Hence 
\begin{equation}
\begin{split}
    \mathcal{AF}_3(u, v) &= 4 - m - n + 3t \leq 4 - m - n + 3(n - 1) \\
    &= 2n - m + 1 \leq n + 1
\end{split}
\end{equation}

\noindent The upper bound is attained when $t = n - 1$ and $n = m$. The upper bound on $\mathcal{AF}_3(u, v)$ follows from the observation that $q \leq (m-1)(n-1)$, so
\begin{equation}
\begin{split}
    \mathcal{AF}_4(u, v) &= 4 - m - n + 3t + 2q \leq 4 - m - n + 3(n - 1) + 2(m-1)(n-1) \\
    &= 2n - m + 1 + 2mn -2n - 2m + 2 = 2mn - 3m + 3 \\
    &\leq 2mn - 3n + 3
\end{split}
\end{equation}

\noindent Note that the bound is attained when $t = n + 1, q = (m-1)(n-1)$, and $m = n$. 

$\hfill \square$ 

\subsubsection{Theorem 3.2}
\label{subsubsection:Thm_3.2}

\noindent \textit{Proof:} $\phi_k$ is $L$-Lipschitz, so
\begin{equation}
\begin{split}
    \left| \mathbf{X}_u^{k + 1} - \mathbf{X}_v^{k + 1}\right| &= \left| \phi_k \left( \bigoplus_{p \in \tilde{N}_u} \psi_k \left( \mathbf{X}_p^k\right)\right) - \phi_k \left( \bigoplus_{q \in \tilde{N}_v} \psi_k \left( \mathbf{X}_q^k\right)\right)\right| \\
    &\leq L \left| \left( \bigoplus_{p \in \tilde{N}_u} \psi_k \left( \mathbf{X}_p^k\right)\right) - \left( \bigoplus_{q \in \tilde{N}_v} \psi_k \left( \mathbf{X}_q^k\right) \right)\right| \\
    &= L \left| \sum_{p \in \tilde{N}_u} \psi_k \left( \mathbf{X}_p^k\right) -  \sum_{q \in \tilde{N}_v} \psi_k \left( \mathbf{X}_q^k\right)\right| \\
    &= L \left| \sum_{p \in \tilde{N}_u \setminus \tilde{N}_v} \psi_k \left( \mathbf{X}_p^k\right) - \sum_{q \in \tilde{N}_v \setminus \tilde{N}_u} \psi_k \left( \mathbf{X}_q^k\right)\right| \leq L \sum_{p \in \tilde{N}_u \Delta \tilde{N}_v} \left|\psi_k \left( \mathbf{X}_p^k\right) \right|
\end{split} 
\end{equation}

\noindent By \textbf{Lemma 1} and using that $t \leq m - 1$ and $q \leq (m-1)(n-1)$, we have after some algebra
\begin{equation}
\begin{split}
    \left| \tilde{N}_u \Delta \tilde{N}_v \right| &\leq n + m - 2t - 2 \leq 3m - 3t - 3 \\
    & \leq \left( 2mn - 2n - 2m + 2 \right) + 3m - 3 - 3t - 2q \\
    &= 2mn - 3n + 3 - \left( 4 - m - n + 3t + 2q \right) \\
    &= 2mn - 3n + 3 - \mathcal{AF}_4(u, v)
\end{split}
\end{equation}

\noindent Finally, using the above and the assumptions that $\psi_k$ is bounded and that $|\mathbf{X}_p^k| \leq C$ for all $p \in \mathcal{N}(u) \cup \mathcal{N}(v)$,
\begin{equation}
    L \sum_{p \in \tilde{N}_u \Delta \tilde{N}_v} \left|\psi_k \left( \mathbf{X}_p^k\right) \right| \leq LCM \left(2mn - 3n + 3 - \mathcal{AF}_4(u, v) \right)
\end{equation}

\noindent This proves the statement with $H := LCM$. 

$\hfill \square$ 

\subsubsection{Proposition 3.3}
\label{subsubsection:Prop_3.3}

\noindent \textit{Proof:} Since $\mathcal{AF}_4(u, v) \in \mathbb{N}$ for all $(u, v) \in E$, we may assume that $\delta \in \mathbb{N}$. We proceed by induction. For all edges $u \sim v$, \textbf{Lemma 1} tells us that
\begin{equation}
\begin{split}
    \left| \tilde{N}_u \Delta \tilde{N}_v \right| &\leq \text{deg}(u) + \text{deg}(v) - 2t - 2 \leq 3m - 3t - 3\\ &\leq 2mn - 3n + 3 - \mathcal{AF}_4(u, v) \leq 2mn - 3n + 3 - \delta
\end{split}
\end{equation}

\noindent The base case $k = 1$ follows, since
\begin{equation}
\begin{split}
    \left| \mathbf{X}_u^{1} - \mathbf{X}_v^{1}\right| &= \left| \phi_1 \left( \frac{1}{n+1}\sum_{p \in \tilde{N}_u} \psi \left( \mathbf{X}_p\right) \right) -  \phi_1 \left( \frac{1}{n+1} \sum_{q \in \tilde{N}_v} \psi \left( \mathbf{X}_q\right) \right)\right| \\
    &\leq L \left| \frac{1}{n+1} \sum_{p \in \tilde{N}_u} \psi \left( \mathbf{X}_p\right) - \frac{1}{n+1} \sum_{q \in \tilde{N}_v} \psi \left( \mathbf{X}_q\right)\right| \\
    &= \frac{L}{n+1}\left| \sum_{p \in \tilde{N}_u \setminus \tilde{N}_v} \psi \left( \mathbf{X}_p\right) -  \sum_{q \in \tilde{N}_v \setminus \tilde{N}_u} \psi \left( \mathbf{X}_q\right)\right| \\
    &\leq \frac{L}{n+1} \sum_{p \in \tilde{N}_u \Delta \tilde{N}_v} \left| \psi \left( \mathbf{X}_p\right) \right| \\
    & \leq \left(\frac{2mn - 3n + 3 - \delta}{n + 1} \right) LCM
\end{split}
\end{equation}

\noindent Suppose now that the statement is true for $k$ and consider the case $k+1$. We have for all $u \sim v$:
\begin{equation}
\begin{split}
    \left| \mathbf{X}_u^{k} - \mathbf{X}_v^{k}\right| &\leq \frac{L}{n+1} \left|  \sum_{p \in \tilde{N}_u} \psi_k \left( \mathbf{X}_p^k\right) - \sum_{q \in \tilde{N}_v} \psi_k \left( \mathbf{X}_q^k\right)\right| \\
    &= \frac{L}{n+1} \left|  \sum_{p \in \tilde{N}_u \setminus \tilde{N}_v} \psi_K \left( \mathbf{X}_p^k\right) - \sum_{q \in \tilde{N}_v \setminus \tilde{N}_u} \psi_k \left( \mathbf{X}_q^k\right)\right| \\
    &= \frac{L}{n+1} \left|  \psi_k \left(\sum_{p \in \tilde{N}_u \setminus \tilde{N}_v} \mathbf{X}_p^k - \sum_{q \in \tilde{N}_v \setminus \tilde{N}_u}  \mathbf{X}_q^k\right)\right| \\
    &\leq \frac{LM}{n+1} \left|  \sum_{p \in \tilde{N}_u \setminus \tilde{N}_v} \mathbf{X}_p^k - \sum_{q \in \tilde{N}_v \setminus \tilde{N}_u}  \mathbf{X}_q^k \right| \\
\end{split}
\end{equation}

\noindent For each $p \in \tilde{N}_u \setminus \tilde{N}_v$, match it with one and only one $q \in \tilde{N}_v \setminus \tilde{N}_u$. For any node pair, they are connected by a node path $p \sim u \sim v \sim q$, where the difference in norm of features at layer $k$ of each $1$-hop connection is at most $\frac{1}{3}C \left(\frac{3LM}{n+1}(2mn - 3n + 3 - \delta)\right)^k$. Hence, we have
\begin{equation}
\begin{split}
    \left| \mathbf{X}_p^{k} - \mathbf{X}_q^{k}\right| \leq C \left(\frac{3LM(2mn - 3n + 3 - \delta)}{n+1}\right)^k
\end{split}
\end{equation}

\noindent Substitute this back into the above and note that there are at most $2mn - 3n + 3 - \delta$ pairs to get
\begin{equation}
\begin{split}
    \left| \mathbf{X}_u^{k+1} - \mathbf{X}_v^{k+1}\right|  &\leq \frac{LM}{n+1} \sum_{(p, q)} \left| \mathbf{X}_p^{k} - \mathbf{X}_q^{k}\right| \\
    & \leq \frac{LM}{n+1} \left(2mn - 3n + 3 - \delta\right) C \left(\frac{3LM(2mn - 3n + 3 - \delta)}{n+1} \right)^k \\
    &= \frac{1}{3}C \left(\frac{3LM(2mn - 3n + 3 - \delta)}{n+1} \right)^{k+1}
\end{split}
\end{equation}

\noindent This proves that the desired inequality holds for all $k \geq 1$ and $u \sim v$. 

$\mathcolor{blue}{\hfill \square}$

\subsubsection{Proposition 3.4}
\label{subsubsection:Prop_3.4}

\noindent \textit{Proof:} Denote the number of triangles containing $(u, v)$ by $\triangle(u,v)=:t$ and the number of quadrangles by $\square (u,v)=:q$. Then $|S| = t + q$. By definition,
\begin{equation}
    \mathcal{AF}_4(u, v) = 4 - \text{deg}(u) - \text{deg}(v) + 3t + 2q = 4 - \text{deg}(u) - \text{deg}(v) + t + 2|S|.
\end{equation}

\noindent Rewriting this, we get
\begin{equation}
    \left| S\right| = \frac{\mathcal{AF}_4(u, v) + \text{deg}(u) + \text{deg}(v) - 4 - t}{2},
\end{equation}

\noindent but $t \geq 0$, so the inequality follows.

$\hfill \square$ 

\subsubsection{Theorem 4.1}
\label{subsubsection:Thm_4.1}

\noindent \textit{Proof:} The complexity of computing discrete curbature in the sense of Ollivier and Forman follows directly from the definitions~\citep{Ol2,forman}; we will briefly recall it below for completeness. Computing ORC involves the computation of the $W_1$-distance between measures defined on the neighborhoods of the nodes adjacent to the edge, which can be done in $O(d_{max}^3)$ (via the Hungarian algorithm), where $d_{\max}$ is the maximal node degree.
For computing $\mathcal{AF}_3$, we note that the costliest operation involved is counting the number of triangles containing a given edge $(u, v) \in E$. We can do this by determining the neighborhoods $\mathcal{N}_u$ and $\mathcal{N}_v$ of $u$ and $v$, respectively, which scales linearly in $\text{max}\left\{d_u, d_v \right\} \leq d_\text{max}$, before taking the intersection of the two neighborhoods, again in linear time. Doing this for every edges shows that computing $\mathcal{AF}_3$ indeed scales as $O\left(\left|E\right| d_\text{max} \right)$. For $\mathcal{AF}_4$, note that counting the 4-cycles containing $(u, v)$ is the costliest operation involved. For a given edge, this scales quadratically in $\text{max}\left\{d_u, d_v \right\} \leq d_\text{max}$: for each $w_1 \in \mathcal{N}_u$ and for each $w_2 \in \mathcal{N}_v \setminus \mathcal{N}_u$, we check whether $(w_1, w_2) \in E$. 

$\hfill \square$

\subsection{Additional theoretical results}

\subsubsection{Lemma 1}
\label{subsubsection:Lemma_1}

\noindent \textbf{Lemma 1:} Let $\tilde{N}_u = N_u \cup \{ u\}$, i.e. $\tilde{N}_u$ contains the neighborhood of the vertex $u$ and $u$ itself. Let $G = (V, E)$ be a graph and suppose that $(u, v) \in E$. Then
\begin{equation}
    \left| \tilde{N}_u \Delta \tilde{N}_v \right| \leq \text{deg}(u) + \text{deg}(v) - 2t - 2
\end{equation}

\noindent \textit{Proof:} Note that $\left| \tilde{N}_u \Delta \tilde{N}_v \right| = \left| \left(\tilde{N}_u \setminus \tilde{N}_v\right) \cup \left(\tilde{N}_v \setminus \tilde{N}_u\right) \right|$, and that for $\left|\tilde{N}_u \setminus \tilde{N}_v\right|$, we have
\begin{equation}
    \left|\tilde{N}_u \setminus \tilde{N}_v\right| = \left| N_u \setminus N_v\right| - 1 = \left| N_u \setminus N_u \cap N_v \right| - 1 = \left| N_u \right| - \left| N_u \cap N_v\right| - 1 = \text{deg}(u) - t - 1
\end{equation}

\noindent The result follows by symmetry in $u$ and $v$ and because 

\begin{equation}
    \left| \left(\tilde{N}_u \setminus \tilde{N}_v\right) \cup \left(\tilde{N}_v \setminus \tilde{N}_u\right) \right| \leq \left| \left(\tilde{N}_u \setminus \tilde{N}_v\right) \right| + \left| \left(\tilde{N}_v \setminus \tilde{N}_u\right) \right|
\end{equation}

$\hfill \square$ 

\subsubsection{Extension of Theorem 3.2}
\label{subsubsection:Thm_3.2_ext}

\noindent \textbf{Theorem (Extension of Theorem 3.2):} Consider an updating rule as defined above and suppose that $\mathcal{AF}_3(u, v) > 0$. For some $k$, assume the update function $\phi_k$ is $L$-Lipschitz, $|\mathbf{X}_p^k| \leq C$ is bounded for all $p \in \mathcal{N}(u) \cup \mathcal{N}(v)$, and the message passing function is bounded, i.e. $|\psi_k(\mathbf{x})| \leq M |\mathbf{x}|, \forall \mathbf{x}$. Let $\bigoplus$ be the mean operation. Then there exists a constant $H > 0$, depending on $L, M$, and $C$, such that 
\begin{equation}
    \left| \mathbf{X}_u^{k + 1} - \mathbf{X}_v^{k + 1}\right| \leq (n + 1 - \mathcal{AF}_3(u, v))h(\mathcal{AF}_3(u, v))
\end{equation}

\noindent where $m = \text{deg}(u) \geq \text{deg}(v) = n > 1$. Furthermore, we have
\begin{equation}
    \lim_{\mathcal{AF}_3(u, v) \rightarrow n+1} (n + 1 - \mathcal{AF}_3(u, v))h(\mathcal{AF}_3(u, v)) = 0
\end{equation}

\noindent \textit{Proof:} We have
\begin{equation}
\begin{split}
    \left| \mathbf{X}_u^{k + 1} - \mathbf{X}_v^{k + 1}\right| &\leq L \left| \left( \bigoplus_{p \in \tilde{N}_u} \psi_k \left( \mathbf{X}_p^k\right)\right) - \left( \bigoplus_{q \in \tilde{N}_v} \psi_k \left( \mathbf{X}_q^k\right) \right)\right| \\
    &= L \left| \sum_{p \in \tilde{N}_u} \frac{1}{n+1} \psi_k \left( \mathbf{X}_p^k\right) -  \sum_{q \in \tilde{N}_v} \frac{1}{m+1} \psi_k \left( \mathbf{X}_q^k\right)\right| \\
    &\leq L \sum_{p \in \left( \tilde{N}_u \cap \tilde{N}_v \right)} \left( \frac{1}{m + 1} - \frac{1}{n+1} \right) \left| \psi_k \left( \mathbf{X}_p^k\right) \right| \\
    &+ L \left| \sum_{p \in \tilde{N}_u \setminus \tilde{N}_v} \frac{1}{n+1} \psi_k \left( \mathbf{X}_p^k\right) - \sum_{q \in \tilde{N}_v \setminus \tilde{N}_u} \frac{1}{m+1} \psi_k \left( \mathbf{X}_q^k\right) \right|
\end{split}
\end{equation}

\noindent Note that $\mathcal{AF}_3(u, v) \leq \text{min} \left \{ m + 1, n + 1\right\}$, so
\begin{equation}
\begin{split}
    \frac{1}{m + 1} - \frac{1}{n + 1} \leq \frac{1}{\mathcal{AF}_3(u, v)} - \frac{1}{n + 1} = \frac{n + 1 - \mathcal{AF}_3(u, v)}{(n+1) \mathcal{AF}_3(u, v)}
\end{split}
\end{equation}

\noindent Hence the above equation now implies

\begin{equation}
\begin{split}
    \left| \mathbf{X}_u^{k + 1} - \mathbf{X}_v^{k + 1}\right| &\leq L \sum_{p \in \left( \tilde{N}_u \cap \tilde{N}_v \right)} \frac{n + 1 - \mathcal{AF}_3(u, v)}{(n+1) \mathcal{AF}_3(u, v)}   
    \left| \psi_k \left( \mathbf{X}_p^k\right) \right|
    + L  \sum_{p \in \tilde{N}_u \Delta \tilde{N}_v} \frac{1}{\mathcal{AF}_3(u, v)} \left| \psi_k \left( \mathbf{X}_p^k\right) \right| \\
    &\leq LCM(n + 1) \frac{n + 1 - \mathcal{AF}_3(u, v)}{(n+1) \mathcal{AF}_3(u, v)} + LCM\left(n + 1 - \mathcal{AF}_3(u, v)\right) \frac{1}{\mathcal{AF}_3(u, v)} \\
    &= 2LCM\frac{n + 1 - \mathcal{AF}_3(u, v)}{\mathcal{AF}_3(u, v)}
\end{split}
\end{equation}

\noindent This proves the statement with $h(\mathcal{AF}_3(u, v)) := \frac{2LCM}{\mathcal{AF}_3(u, v)}$.

$\hfill \square$ 

\subsubsection{Extension of Proposition 3.3}
\label{subsubsection:Prop_3.3_ext}

\noindent \textbf{Theorem (Extension of Proposition 3.3):} Consider an updating rule as defined above. Assume the graph is regular. Suppose there exists a constant $\delta$, such that for all edges $(u, v) \in E$, $\mathcal{AF}_3(u, v) \geq \delta > 0$. For all $k$, assume the update functions $\phi_k$ are $L$-Lipschitz, $\bigoplus$ is the mean operation, $|\mathbf{X}_p^0| \leq C$ is bounded for all $p \in V$, and the message passing functions are bounded linear operators, i.e. $|\psi_k(\mathbf{x})| \leq M |\mathbf{x}|, \forall \mathbf{x}$. Then the following inequality holds for $k \geq 1$ and any neighboring vertices $u \sim v$
\begin{equation}
    \left| \mathbf{X}_u^{k} - \mathbf{X}_v^{k}\right| \leq \frac{1}{3}C \left(\frac{3LM(n + 1 -\delta)}{n+1} \right)^k
\end{equation}

\noindent \textit{Proof:} Since $\mathcal{AF}_3(u, v) \in \mathbb{N}$ for all $(u, v) \in E$, we may assume that $\delta \in \mathbb{N}$. We proceed by induction. For all edges $u \sim v$, \textbf{Lemma 1} tells us that 
\begin{equation}
\begin{split}
    \left| \tilde{N}_u \Delta \tilde{N}_v \right| &\leq \text{deg}(u) + \text{deg}(v) - 2t - 2\\
    &\leq n + 1 - \mathcal{AF}_3(u, v) \leq n + 1 - \delta
\end{split}
\end{equation}

\noindent The base case $k = 1$ follows, since
\begin{equation}
\begin{split}
    \left| \mathbf{X}_u^{1} - \mathbf{X}_v^{1}\right| 
    &\leq L \left| \frac{1}{n + 1}\sum_{p \in \tilde{N}_u}  \psi \left( \mathbf{X}_p\right) -  \frac{1}{n + 1} \sum_{q \in \tilde{N}_v} \psi \left( \mathbf{X}_q\right)\right| \\
    &= \frac{L}{n + 1}\left| \sum_{p \in \tilde{N}_u \setminus \tilde{N}_v} \psi \left( \mathbf{X}_p\right) -  \sum_{q \in \tilde{N}_v \setminus \tilde{N}_u} \psi \left( \mathbf{X}_q\right)\right| \\
    &\leq \frac{L}{n + 1} \sum_{p \in \tilde{N}_u \Delta \tilde{N}_v} \left| \psi \left( \mathbf{X}_p\right) \right| \\
    & \leq \frac{(n + 1 - \delta)}{n + 1} LCM
\end{split}
\end{equation}

\noindent Suppose now that the statement is true for $k$ and consider the case $k+1$. We have for all $u \sim v$:
\begin{equation}
\begin{split}
    \left| \mathbf{X}_u^{k} - \mathbf{X}_v^{k}\right| &\leq \frac{L}{n + 1} \left|  \sum_{p \in \tilde{N}_u} \psi_k \left( \mathbf{X}_p^k\right) - \sum_{q \in \tilde{N}_v} \psi_k \left( \mathbf{X}_q^k\right)\right| \\
    &= \frac{L}{n + 1} \left|  \sum_{p \in \tilde{N}_u \setminus \tilde{N}_v} \psi_K \left( \mathbf{X}_p^k\right) - \sum_{q \in \tilde{N}_v \setminus \tilde{N}_u} \psi_k \left( \mathbf{X}_q^k\right)\right| \\
    &= \frac{L}{n + 1} \left|  \psi_k \left(\sum_{p \in \tilde{N}_u \setminus \tilde{N}_v} \mathbf{X}_p^k - \sum_{q \in \tilde{N}_v \setminus \tilde{N}_u}  \mathbf{X}_q^k\right)\right| \\
    &\leq \frac{LM}{n + 1} \left|  \sum_{p \in \tilde{N}_u \setminus \tilde{N}_v} \mathbf{X}_p^k - \sum_{q \in \tilde{N}_v \setminus \tilde{N}_u}  \mathbf{X}_q^k \right| \\
\end{split}
\end{equation}

\noindent For each $p \in \tilde{N}_u \setminus \tilde{N}_v$, match it with one and only one $q \in \tilde{N}_v \setminus \tilde{N}_u$. For any node pair, they are connected by a node path $p \sim u \sim v \sim q$, where the difference in norm of features at layer $k$ of each $1$-hop connection is at most $\frac{1}{3}C \left(\frac{3LM(n + 1 - \delta)}{n + 1}\right)^k$. Hence, we have
\begin{equation}
\begin{split}
    \left| \mathbf{X}_p^{k} - \mathbf{X}_q^{k}\right| \leq C \left(\frac{3LM(n + 1 - \delta)}{n + 1}\right)^k
\end{split}
\end{equation}

\noindent Substitute this back into the above and note that there are at most $n + 1 - \delta$ pairs to get
\begin{equation}
\begin{split}
    \left| \mathbf{X}_u^{k+1} - \mathbf{X}_v^{k+1}\right|  &\leq \frac{LM}{n + 1} \sum_{(p, q)} \left| \mathbf{X}_p^{k} - \mathbf{X}_q^{k}\right| \\
    & \leq \frac{LM}{n + 1} \left(n + 1 - \delta(u, v)\right) C \left(3LM(n + 1 - \delta) \right)^k \\
    &= \frac{1}{3}C \left(\frac{3LM(n + 1 - \delta)}{n + 1} \right)^{k+1}
\end{split}
\end{equation}

\noindent This proves that the desired inequality holds for all $k \geq 1$ and $u \sim v$. 

$\hfill \square$ 

\subsubsection{Theorem 4.5 in Nguyen et al. (2023)}

\noindent The following result from (Nguyen et al. 2023) also applies when using augmentations of the Forman-Ricci curvature. It shows that edges with $\mathcal{AF}_4$ close to the lower bound result in a decaying importance of distant nodes in GNNs without non-linearities. 

\noindent \textbf{Theorem 4.5.} Consider the same updating rule as before, let $\psi_k$ and $\phi_k$ be linear operators for all $k$, and let $\bigoplus$ be the sum operation. Let $u$ and $v$ be neighboring vertices with neighborhoods as in the previous theorem and let $S$ be defined similarly. Then for all $p \in \tilde{N}_u \setminus \{ v\}$ and $q \in \tilde{N}_v \setminus \{ u\}$, we have
\begin{equation}
\begin{split}
    \left[ \frac{\delta \mathbf{X}_u^{k + 2}}{\delta \mathbf{X}_q^k}\right] = \alpha \sum_{w \in V} \left[ \frac{\delta \mathbf{X}_u^{k + 2}}{\delta \mathbf{X}_w^k} \right]; \qquad
    \left[ \frac{\delta \mathbf{X}_v^{k + 2}}{\delta \mathbf{X}_p^k} \right] = \beta \sum_{w \in V} \left[ \frac{\delta \mathbf{X}_v^{k + 2}}{\delta \mathbf{X}_w^k}\right]
\end{split}
\end{equation}

\noindent where $\left[ \frac{\mathbf{y}}{\mathbf{x}}\right]$ denotes the Jacobian of $\mathbf{y}$ w.r.t. $\mathbf{x}$, and $\alpha$ and $\beta$ satisfy
\begin{equation}
\begin{split}
    \alpha \leq \frac{|S| + 2}{\sum_{w \in \tilde{N}_v} \left(\text{deg}(w) + 1\right)}; \qquad
    \beta \leq \frac{|S| + 2}{\sum_{w \in \tilde{N}_u} \left(\text{deg}(w) + 1\right)}
\end{split}
\end{equation}

\noindent For the proof, see \cite{nguyen2023revisiting}.

\subsection{Best hyperparameter settings}
\label{subsubsection:hyperparam}

\noindent Our hyperparameter choices are largely based on the hyperparameters reported for BORF in \cite{nguyen2023revisiting}. We used their values as starting values for a grid search, but found the changes in accuracy to be minimal across datasets. We similarly experimented with different hyperparameter choices for AFR-3, AFR-4, and BORF, but again found the resulting differences in accuracy to be marginal. We therefore decided to use the hyperparameters reported here for AFR-3, AFR-4, and BORF.

\begin{table}[h!]
\centering
\begin{tabular}{|l|c|c|c|c|c|c|}
\hline
 & 
\multicolumn{3}{|c|}{GCN} & \multicolumn{3}{|c|}{GIN} \\
\hline
DATA SET & $n$ & $h$ & $l$ & $n$ & $h$ & $l$ \\
\hline
CORA & 3 & 20 & 10 & 3 & 20 & 30 \\
CITESEER & 3 & 20 & 10 & 3 & 10 & 20 \\
TEXAS & 3 & 30 & 10 & 3 & 20 & 10 \\
CORNELL & 2 & 20 & 30 & 3 & 10 & 20 \\
WISCONSIN & 2 & 30 & 20 & 2 & 50 & 30 \\
CHAMELEON & 3 & 20 & 20 & 3 & 30 & 30 \\
\hline
MUTAG & 1 & 20 & 3 & 1 & 3 & 1 \\
ENZYMES & 1 & 3 & 2 & 3 & 3 & 1 \\
IMDB & 1 & 3 & 0 & 1 & 4 & 2 \\
PROTEINS & 3 & 4 & 1 & 2 & 4 & 3 \\
\hline
\end{tabular}
\label{tab:afrc_orc_hyperparam}
\caption{Best hyperparameter settings for the datasets and architectures considered in this paper. Here, $n$ is the number of iterations, $h$ is the number of edges added, and $l$ is the number of edges removed.}
\end{table}

\subsection{Other rewiring algorithms}
\label{subsubsection:other_algos}

\noindent We compare AFR against three recently introduced rewiring algorithms for GNNs, two of which are based on discrete curvature. SDRF and FoSR were designed with to deal with over-squashing in GNNs, while BORF can address both over-squashing and over-smoothing. BORF is based on the Ollivier-Ricci curvature, while SDRF uses a lower bound for the ORC, called the Balanced Forman curvature (BFR). SDRF iteratively identifies the edge with the lowest BFC, computes the increase in BFC for every possible edge that could be added to the graph, and then adds the edge that increases the original edge's BFR the most. FoSR, on the other hand, is based on the spectral gap, which characterizes the connectivity of a graph. At each step, it approximately identifies the edge that, if added, would maximally increase the spectral gap, and adds that edge to the graph. For SDRF and FoSR, we use the hyperparameters reported in \cite{nguyen2023revisiting}. 

\noindent BORF, as mentioned before, uses the ORC to identify edges responsible for over-smoothing and over-squashing. The algorithm itself is nearly identical to the AFR algorithm without heuristics.

\subsection{Rewiring Times}

\begin{table}[h!]\label{tab:rewiring_times}
\centering
\begin{tabular}{|l|c|c|c|}
\hline
DATA SET & AFR-3 & AFR-4 & BORF \\
\hline
CHAMELEON & 0.3624 & 4.7628 & 81.6509 \\
AMAZON RATINGS & 0.0127 & 0.0487 & 0.1437 \\
MINESWEEPER & 0.0098 & 0.0363 & 0.1415 \\
TOLOKERS & 7.0826 & 95.2717 & Timeout \\
\hline
\end{tabular}

\caption{Time (in seconds) required for rewiring using AFR-3, AFR-4, and BORF. On the Tolokers dataset, BORF does not terminate within 60 minutes.}
\end{table}

\subsection{Architecture choices}

\noindent \textbf{Node classification.} We use a 3-layer GNN with hidden dimension 128, dropout probability 0.5, and ReLU activation. We use this architecture for all node classification datasets considered.

\noindent \textbf{Graph classification.} We use a 4-layer GNN with hidden dimension 64, dropout probability 0.5, and ReLU activation. We use this architecture for all graph classification datasets considered.

\subsection{Additional experimental results}

\subsubsection{Node and graph classification using GIN}
\label{subsubsection:GIN_node_graph_heur}

\begin{table}[h!]\label{tab:afrc_orc_hyperparam}
\centering
\begin{tabular}{|l|c|c|c|c|c|c|}
\hline
 & 
\multicolumn{6}{|c|}{GIN} \\
\hline
DATA SET & AFR-3 & AFR-4 & BORF & SDRF & FOSR & NONE \\
\hline
CORA & $\mathbf{79.2 \pm 0.7}$ & $78.3 \pm 0.6$ & $78.4 \pm 0.4$ & $74.8 \pm 0.3$ & $75.7 \pm 0.9 $ & $76.3 \pm 0.6$ \\
CITESEER & $64.3 \pm 0.9$ & $\mathbf{64.8 \pm 0.7}$ & $63.1 \pm 0.7$ & $60.6 \pm 0.8$ & $61.6 \pm 0.7$ & $59.9 \pm 0.6$ \\
TEXAS & $62.8 \pm 1.8$ & $62.1 \pm 1.6$ & $\mathbf{63.1 \pm 1.6}$ & $50.2 \pm 3.4$ & $47.2 \pm 3.6$ & $53.8 \pm 3.0$ \\
CORNELL & $\mathbf{49.3 \pm 1.1}$ & $48.8 \pm 1.2$ & $48.9 \pm 1.2$ & $40.4 \pm 2.0$ & $36.1 \pm 2.4$ & $36.4 \pm 2.3$ \\
WISCON. & $54.1 \pm 1.3$ & $53.9 \pm 1.2$ & $\mathbf{54.4 \pm 1.3}$ & $48.6 \pm 1.8$ & $48.8 \pm 2.4$ & $48.5 \pm 2.1$\\
CHAMEL. & $\mathbf{66.4 \pm 1.0}$ & $65.9 \pm 0.9$ & $65.5 \pm 0.7$ & $58.7 \pm 2.3$ & $57.1 \pm 2.2$ & $58.8 \pm 2.3$ \\
\hline
COCO & $12.9 \pm 1.8$ & $13.1 \pm 1.9$ & $\mathbf{13.2 \pm 1.8}$ & $9.3 \pm 1.6$ & $10.2 \pm 1.5$ & $8.8 \pm 1.5$ \\
PASCAL & $16.2 \pm 2.0$ & $15.9 \pm 1.9$ & $\mathbf{16.3 \pm 2.1}$ & $12.2 \pm 1.4$ & $14.4 \pm 1.7$ & $12.3 \pm 1.1$ \\
\hline
\end{tabular}

\caption{Classification accuracies of GIN with AFR-3, AFR-4, BORF, SDRF, FoSR, or no rewiring strategy using best hyperparameters. Highest accuracies on any given dataset are highlighted in bold.}
\label{table:1}
\end{table}

\begin{table}[h!]\label{tab:afrc_orc_hyperparam}
\centering
\begin{tabular}{|l|c|c|c|c|c|c|}
\hline
 & 
\multicolumn{6}{|c|}{GIN}\\
\hline
DATA SET & AFR-3 & AFR-4 & BORF & SDRF & FOSR & NONE \\
\hline
MUTAG & $68.8 \pm 3.2$ & $69.3 \pm 2.9$ & $\mathbf{72.1 \pm 3.1}$ & $68.1 \pm 1.1$ & $67.2 \pm 2.9$ & $67.5 \pm 2.7$ \\
ENZYMES & $\mathbf{34.6 \pm 1.4}$ & $33.1 \pm 1.2$ & $33.2 \pm 17.$ & $31.5 \pm 1.3$ & $24.9 \pm 1.4$ & $29.7 \pm 1.1$\\
IMDB & $68.8 \pm 1.4$ & $\mathbf{69.3 \pm 1.5}$ & $68.7 \pm 1.2$ & $66.6 \pm 1.4$ & $67.3 \pm 1.2$ & $67.1 \pm 1.3$ \\
PROTEINS & $71.9 \pm 0.9$ & $\mathbf{72.7 \pm 0.7}$ & $71.2 \pm 0.9$ & $72.1 \pm 0.8$ & $71.8 \pm 0.7$ & $69.4 \pm 1.1$ \\
\hline
PEPTIDES & $49.4 \pm 1.6$ & $50.2 \pm 1.7$ & $49.9 \pm 1.6$ & $46.4 \pm 1.5$ & $48.7 \pm 1.9$ & $46.0 \pm 2.3$ \\
\hline
\end{tabular}

\caption{Classification accuracies of GIN with AFR-3, AFR-4, BORF, SDRF, FoSR, or no rewiring strategy using best hyperparameters. Highest accuracies on any given dataset are highlighted in bold.}
\label{table:1}
\end{table}

\newpage

\subsubsection{Ablations on heuristic for adding edges}
\label{subsubsection:adding_ablations}

\begin{table}[h!]\label{tab:afrc_orc_hyperparam}
\centering
\begin{tabular}{|l|c|c|c|c|c|c|}
\hline
 & 
\multicolumn{3}{|c|}{GCN} & \multicolumn{3}{|c|}{GIN} \\
\hline
DATA SET & AFR-3 & AFR-4 & BORF & AFR-3 & AFR-4 & BORF\\
\hline
CORA & $87.4 \pm 0.7$ & $87.1 \pm 0.6$ & $86.2 \pm 0.7$ & $77.9 \pm 1.3$ & $78.1 \pm 1.2$ & $77.0 \pm 1.0$\\
CITESEER & $73.2 \pm 0.7$ & $73.0 \pm 0.8$ & $73.1 \pm 0.6$ & $64.2 \pm 1.0$ & $63.9 \pm 1.7$ & $63.0 \pm 1.7$\\
TEXAS & $51.9 \pm 4.2$ & $50.0 \pm 3.1$ & $51.9 \pm 4.0$ & $62.6 \pm 1.8$ & $62.3 \pm 1.8$ & $ 62.2 \pm 1.8$\\
CORNELL & $47.8 \pm 4.4$ & $47.3 \pm 4.3$ & $47.8 \pm 3.9$ & $49.1 \pm 1.3$ & $49.0 \pm 1.2$ & $48.5 \pm 1.3$\\
WISCON. & $53.5 \pm 2.2$ & $51.9 \pm 4.2$ & $50.9 \pm 2.5$ & $52.6 \pm 1.1$ & $53.2 \pm 1.9$ & $53.3 \pm 1.3$\\
CHAMEL. & $60.4 \pm 1.1$ & $60.2 \pm 1.0$ & $57.1 \pm 1.4$ & $65.3 \pm 0.8$ & $65.1 \pm 0.9$ & $64.9 \pm 0.7$\\
\hline
COCO & $9.8 \pm 1.1$ & $9.9 \pm 1.2$ & $10.2 \pm 1.1$ & $13.0 \pm 1.9$ & $13.3 \pm 2.1$ & $13.2 \pm 2.1$ \\
PASCAL & $14.4 \pm 1.4$ & $14.1 \pm 1.2$ & $13.7 \pm 1.1$ & $15.8 \pm 1.6$ & $16.1 \pm 1.5$ & $15.9 \pm 1.5$\\
\hline
\end{tabular}

\caption{Classification accuracies of GCN and GIN with AFR-3, AFR-4, and BORF. Here, we only use our heuristic for adding edges and use best hyperparameters to remove edges.}
\label{table:1}
\end{table}

\begin{table}[h!]\label{tab:afrc_orc_heuristic}
\centering
\begin{tabular}{|l|c|c|c|c|c|c|}
\hline
 & 
\multicolumn{3}{|c|}{GCN} & \multicolumn{3}{|c|}{GIN} \\
\hline
DATA SET & AFR-3 & AFR-4 & BORF & AFR-3 & AFR-4 & BORF\\
\hline
MUTAG & $70.8 \pm 2.0$ & $67.6 \pm 2.3$ & $66.7 \pm 1.9$ & $73.1 \pm 2.9$ & $72.3 \pm 2.7$ & $74.7 \pm 2.4$ \\
ENZYMES & $24.3 \pm 1.2$ & $24.5 \pm 1.2$ & $23.9 \pm 1.1$ & $36.3 \pm 1.1$ & $37.2 \pm 1.5$ & $32.3 \pm 1.3$\\
IMDB & $49.2 \pm 0.8$ & $49.1 \pm 1.0$ & $49.1 \pm 1.0$ & $69.4 \pm 0.9$ & $68.1 \pm 0.9$ & $68.1 \pm 1.0$\\
PROTEINS & $59.7 \pm 1.0$ & $59.4 \pm 0.9$ & $59.5 \pm 1.0$ & $74.3 \pm 0.9$ & $72.6 \pm 1.0$ & $70.7 \pm 1.1$ \\
\hline
PEPTIDES & $45.1 \pm 2.7$ & $43.8 \pm 2.6$ & $44.8 \pm 2.6$ & $49.8 \pm 1.4$ & $50.0 \pm 1.4$ & $49.1 \pm 1.6$ \\
\hline
\end{tabular}

\caption{Classification accuracies of GCN and GIN with AFR-3, AFR-4, and BORF. Here, we only use our heuristic for adding edges and use best hyperparameters to remove edges.}
\label{table:2}
\end{table}

\newpage

\subsubsection{Ablations on heuristic for removing edges}
\label{subsubsection:removing_ablations}

\begin{table}[h!]\label{tab:afrc_orc_hyperparam}
\centering
\begin{tabular}{|l|c|c|c|c|c|c|}
\hline
 & 
\multicolumn{3}{|c|}{GCN} & \multicolumn{3}{|c|}{GIN} \\
\hline
DATA SET & AFR-3 & AFR-4 & BORF & AFR-3 & AFR-4 & BORF\\
\hline
CORA & $87.7 \pm 0.5$ & $88.0 \pm 0.5$ & $87.8 \pm 0.5$ & $78.6 \pm 1.3$ & $78.2 \pm 0.9$ & $78.5 \pm 1.6$ \\
CITESEER & $73.8 \pm 1.1$ & $74.3 \pm 0.7$ & $73.7 \pm 1.0$ & $63.5 \pm 1.1$ & $62.7 \pm 0.9$ & $64.2 \pm 0.6$ \\
TEXAS & $50.5 \pm 3.3$ & $48.4 \pm 4.3$ & $52.2 \pm 4.5$ & $63.2 \pm 2.3$ & $65.1 \pm 2.3$ & $58.4 \pm 3.0$ \\
CORNELL & $48.4 \pm 3.4$ & $45.7 \pm 3.1$ & $48.9 \pm 3.3$ & $48.1 \pm 3.9$ & $51.9 \pm 3.8$ & $48.9 \pm 4.1$ \\
WISCON. & $50.2 \pm 3.2$ & $50.6 \pm 3.4$ & $50.6 \pm 3.1$ & $54.3 \pm 1.9$ & $56.9 \pm 2.8$ & $55.9 \pm 2.9$ \\
CHAMEL. & $62.5 \pm 1.1$ & $61.7 \pm 1.0$ & $61.2 \pm 1.0$ & $67.1 \pm 1.4$ & $65.8 \pm 1.4$& $66.0 \pm 1.6$ \\
\hline
COCO & $10.4 \pm 1.2$ & $10.2 \pm 1.1$ & $10.6 \pm 1.4$ & $13.5 \pm 2.3$ & $13.1 \pm 2.2$ & $13.4 \pm 2.3$\\
PASCAL & $14.0 \pm 1.5$ & $14.2 \pm 1.4$ & $14.7 \pm 1.2$ & $16.7 \pm 1.8$ & $ 16.1 \pm 1.6$ & $16.2 \pm 1.5$ \\
\hline
\end{tabular}

\caption{Classification accuracies of GCN and GIN with AFR-3, AFR-4, and BORF. Here, we only use our heuristic for removing edges and use best hyperparameters to add edges.}
\label{table:1}
\end{table}

\begin{table}[h!]\label{tab:afrc_orc_heuristic}
\centering
\begin{tabular}{|l|c|c|c|c|c|c|}
\hline
 & 
\multicolumn{3}{|c|}{GCN} & \multicolumn{3}{|c|}{GIN} \\
\hline
DATA SET & AFR-3 & AFR-4 & BORF & AFR-3 & AFR-4 & BORF\\
\hline
MUTAG & $71.3 \pm 2.1$ & $72.9 \pm 2.2$ & $70.7 \pm 2.3$ & $66.6 \pm 2.8$ & $68.5 \pm 2.6$ & $68.2 \pm 2.7$ \\
ENZYMES & $25.2 \pm 1.1$ & $25.8 \pm 1.2$ & $25.0 \pm 1.2$ & $33.2 \pm 1.3$ & $34.4 \pm 1.3$ & $33.9 \pm 1.4$ \\
IMDB & $48.7 \pm 1.0$ & $49.2 \pm 1.0$ & $49.2 \pm 0.9$ & $69.0 \pm 1.3$ & $69.5 \pm 1.4$ & $68.3 \pm 1.6$ \\
PROTEINS & $60.1 \pm 0.9$ & $59.1 \pm 0.9$ & $59.0 \pm 0.9$ & $72.3 \pm 1.2$ & $70.8 \pm 1.3$ & $71.1 \pm 1.3$ \\
\hline
PEPTIDES & $44.5 \pm 2.6$ & $44.0 \pm 2.5$ & $44.7 \pm 2.6$ & $50.3 \pm 1.6$ & $49.7 \pm 1.5$ & $49.9 \pm 1.6$\\
\hline
\end{tabular}

\caption{Classification accuracies of GCN and GIN with AFR-3, AFR-4, and BORF. Here, we only use our heuristic for removing edges and use best hyperparameters to add edges.}
\label{table:2}
\end{table}

\subsubsection{Multiple rewiring iterations}
\label{subsubsection:multiple_rewiring}

\begin{table}[h!]\label{tab:afrc_orc_heuristic}
\centering
\begin{tabular}{|l|c|c|c|c|c|c|c|c|}
\hline
 & 
\multicolumn{3}{|c|}{GCN} & \multicolumn{3}{|c|}{GIN} \\
\hline
DATA SET & AFR-3 & AFR-4 & BORF & AFR-3 & AFR-4 & BORF\\
\hline
MUTAG & $67.6 \pm 1.9$ & $67.0 \pm 2.0 $ & $67.5 \pm 1.9$ & $72.7 \pm 3.6$ & $73.9 \pm 2.8$ & $76.6 \pm 2.7$ \\
ENZYMES & $27.6 \pm 1.2$ & $26.1 \pm 1.3$ & $32.5 \pm 1.3$ & $36.7 \pm 1.5$ & $37.6 \pm 1.2$ & $36.3 \pm 1.3$ \\
IMDB & $48.5 \pm 0.9$ & $49.0 \pm 0.9$ & $48.7 \pm 0.9$ & $50.2 \pm 1.0$ & $49.7 \pm 0.8$ & $49.5 \pm 0.9$ \\
PROTEINS & $59.4 \pm 0.8$ & $59.1 \pm 0.9$ & $59.9 \pm 0.8$ & $71.7 \pm 1.3$ & $70.2 \pm 1.2$ & $70.9 \pm 1.4$ \\
\hline
PEPTIDES & $43.6 \pm 2.5$ & $44.2 \pm 2.4$ & $44.4 \pm 2.4$ & $49.6 \pm 1.5$ & $50.2 \pm 1.3$ & $50.5 \pm 1.3$\\
\hline
\end{tabular}

\caption{Comparison using our heuristics for edge addition and removal with two iterations.}
\label{table:3}
\end{table}

\begin{table}[h!]\label{tab:afrc_orc_heuristic}
\centering
\begin{tabular}{|l|c|c|c|c|c|c|c|c|}
\hline
 & 
\multicolumn{3}{|c|}{GCN} & \multicolumn{3}{|c|}{GIN} \\
\hline
DATA SET & AFR-3 & AFR-4 & BORF & AFR-3 & AFR-4 & BORF\\
\hline
MUTAG & $67.1 \pm 1.7$ & $67.4 \pm 1.8$ & $71.5 \pm 2.3$ & $72.3 \pm 3.4$ & $73.8 \pm 3.0$ & $74.6 \pm 2.3$ \\
ENZYMES & $22.8 \pm 1.1$ & $22.0 \pm 1.3$ & $21.3 \pm 1.1$ & $31.1 \pm 1.2$ & $30.8 \pm 1.2$ & $30.2 \pm 1.1$ \\
IMDB & $48.0 \pm 0.8$ & $47.7 \pm 0.9$ & $48.6 \pm 1.0$ & $48.7 \pm 1.3$ & $49.1 \pm 1.2$ & $49.0 \pm 1.0$ \\
PROTEINS & $58.3 \pm 0.6$ & $59.0 \pm 0.9$ & $59.6 \pm 0.8$ & $70.1 \pm 1.1$ & $69.8 \pm 1.0$ & $69.2 \pm 1.2$ \\
\hline
PEPTIDES & $43.7 \pm 2.4$ & $43.8 \pm 2.6$ & $44.9 \pm 2.7$ & $50.6 \pm 1.6$ & $50.4 \pm 1.6$ & $49.7 \pm 1.4$\\
\hline
\end{tabular}

\caption{Comparison using our heuristics with a variable number of iterations. We continue until no edges with a curvature below the threshold $\Delta_L$ ($0$ in the case of the ORC) or above the threshold $\Delta_U$ are left.}
\label{table:4}
\end{table}

\newpage

\subsubsection{Effects of GNN depth}
\label{subsubsection:depth}

\begin{table}[h!]
\centering
\begin{tabular}{|l|c|c|c|c|c|}
\hline
 & 
\multicolumn{5}{|c|}{GCN} \\
\hline
DATA SET & AFR-3 & AFR-4 & BORF & DROPEDGE & NONE \\
\hline
CORA & $88.63 \pm 0.67$ & $87.90 \pm 0.82$ & $\mathbf{88.89 \pm 0.67}$ & $86.15 \pm 0.79$ & $84.43 \pm 0.61$ \\
CITESEER & $\mathbf{78.72 \pm 0.48}$ & $78.11 \pm 0.56$ & $78.59 \pm 0.51$ & $78.19 \pm 0.62$ & $75.62 \pm 0.48$ \\
\hline
\end{tabular}

\caption{Classification accuracies of GCN with 4 layers. Highest accuracies highlighted in bold.} 
\end{table}

\begin{table}[h!]
\centering
\begin{tabular}{|l|c|c|c|c|c|}
\hline
 & 
\multicolumn{5}{|c|}{GCN} \\
\hline
DATA SET & AFR-3 & AFR-4 & BORF & DROPEDGE & NONE \\
\hline
CORA & $83.98 \pm 0.51$ & $84.13 \pm 0.62$ & $84.21 \pm 0.55$ & $\mathbf{84.36 \pm 0.72}$ & $77.91 \pm 0.43$\\
CITESEER & $\mathbf{77.73 \pm 0.92}$ & $76.44 \pm 1.18$ & $77.58 \pm 1.06$ & $76.27 \pm 1.24$ & $73.85 \pm 0.87$\\
\hline
\end{tabular}

\caption{Classification accuracies of GCN with 8 layers. Highest accuracies highlighted in bold.} 
\end{table}

\begin{table}[h!]
\centering
\begin{tabular}{|l|c|c|c|c|c|}
\hline
 & 
\multicolumn{5}{|c|}{GCN} \\
\hline
DATA SET & AFR-3 & AFR-4 & BORF & DROPEDGE & NONE \\
\hline
CORA & $83.87 \pm 0.50$ & $84.22 \pm 0.58$ & $\mathbf{84.86 \pm 0.43}$ & $81.59 \pm 0.63$ & $81.54 \pm 0.63$ \\
CITESEER & $\mathbf{72.71 \pm 0.52}$ & $72.64 \pm 0.66$ & $72.14 \pm 0.54$ & $71.49 \pm 0.70$ & $67.36 \pm 0.49$ \\
\hline
\end{tabular}

\caption{Classification accuracies of GCN with 16 layers. Highest accuracies highlighted in bold.} 
\end{table}

\begin{table}[h!]
\centering
\begin{tabular}{|l|c|c|c|c|c|}
\hline
 & 
\multicolumn{5}{|c|}{GCN} \\
\hline
DATA SET & AFR-3 & AFR-4 & BORF & DROPEDGE & NONE \\
\hline
CORA & $75.79 \pm 1.30$ & $75.13 \pm 1.44$ & $\mathbf{76.47 \pm 2.04}$ & $74.62 \pm 1.17$ & $71.31 \pm 0.84$ \\
CITESEER & $65.60 \pm 0.67$ & $\mathbf{66.05 \pm 0.73}$ & $65.84 \pm 0.64$ & $62.93 \pm 1.20$ & $59.82 \pm 0.78$\\
\hline
\end{tabular}

\caption{Classification accuracies of GCN with 32 layers. Highest accuracies highlighted in bold.} 
\end{table}

\begin{table}[h!]\label{tab:afrc_orc_hyperparam}
\centering
\begin{tabular}{|l c|c|c|c|c|}
\hline
 & & \multicolumn{3}{|c|}{HEURISTIC} & \\
\hline
DATA SET & \# LAYERS & AFR-3 & AFR-4 & BORF & NONE \\
\hline

& 4 & $71.4 \pm 2.2$ & $69.9 \pm 2.6$ & $68.5 \pm 1.9 $ & $62.7 \pm 2.1$\\
MUTAG & 6 & $68.1 \pm 2.1$ & $67.3 \pm 2.3$ & $67.4 \pm 2.0$ & $61.2 \pm 1.6$ \\
& 8 & $64.7 \pm 1.6$ & $65.4 \pm 1.4$ & $65.1 \pm 1.4$ & $58.6 \pm 2.2$\\
& 10 & $63.8 \pm 1.5$ & $63.1 \pm 1.5$ & $63.7 \pm 1.6$ & $57.5 \pm 1.4$\\
\hline

& 4 & $26.1 \pm 1.0$ & $25.5 \pm 1.0$ & $23.3 \pm 1.2$ & $25.4 \pm 1.3$\\
ENZYMES & 6 & $25.5 \pm 1.2$ & $24.8 \pm 1.1$ & $25.1 \pm 1.1$ & $24.6 \pm 1.3$\\
& 8 & $23.1 \pm 1.1$ & $22.8 \pm 0.9$ & $23.0 \pm 1.1$ & $22.2 \pm 1.4$\\
& 10 & $21.3 \pm 1.1$ & $20.8 \pm 1.2$ & $20.9 \pm 1.2$ & $20.4 \pm 0.8$\\
\hline

& 4 & $50.1 \pm 0.9$ & $49.0 \pm 0.9$ & $49.4 \pm 1.0$ & $48.1 \pm 1.0$\\
IMDB & 6 & $49.0 \pm 1.2$ & $49.2 \pm 1.2$ & $49.8 \pm 1.3$ & $47.6 \pm 1.1$\\
& 8 & $46.4 \pm 1.2$ & $46.1 \pm 1.0$ & $46.0 \pm 1.2$ & $44.9 \pm 1.4$\\
& 10 & $41.7 \pm 1.2$ & $42.5 \pm 1.4$& $42.1 \pm 1.1$ & $39.8 \pm 1.2$\\
\hline

& 4 & $62.2 \pm 0.8$ & $61.2 \pm 0.9$ & $61.0 \pm 0.9$ & $59.6 \pm 0.9$\\
PROTEINS & 6 & $60.8 \pm 1.3$ & $60.0 \pm 1.0$ & $60.2 \pm 1.0$ & $59.3 \pm 0.8$\\
& 8 & $58.8 \pm 1.1$ & $58.1 \pm 0.7$ & $58.9 \pm 0.9$ & $57.0 \pm 1.1$\\
& 10 & $56.4 \pm 0.8$ & $55.7 \pm 1.1$& $56.3 \pm 0.9$ & $54.5 \pm 1.2$\\
\hline

& 4 & $44.8 \pm 2.8$ & $43.6 \pm 2.5$ & $44.3 \pm 2.8$ & $40.5 \pm 2.1$ \\
PEPTIDES & 6 & $47.2 \pm 2.6$ & $46.1 \pm 2.5$ & $47.4 \pm 2.6$ & $44.0 \pm 2.3$ \\
& 8 & $49.5 \pm 2.5$ & $49.3 \pm 2.7$ & $49.8 \pm 2.5$ & $46.6 \pm 2.4$ \\
& 10 & $50.3 \pm 2.6$ & $50.5 \pm 2.6$ & $50.6 \pm 2.2$ & $48.7 \pm 2.5$\\
\hline
\end{tabular}

\caption{Graph classification accuracy using heuristics for edge addition and removal with increasing GNN depth.}
\label{table:1}
\end{table}

\subsubsection{HeterophilousGraphDataset}

\begin{table}[h!]
\centering
\tiny
\begin{tabular}{|l|c|c|c|c|c|c|}
\hline
 & 
\multicolumn{6}{|c|}{GCN} \\
\hline
DATA SET & AFR-3 & AFR-4 & BORF & SDRF & FOSR & NONE \\
\hline
AMAZON RATINGS & $46.91 \pm 0.40$ & $47.14 \pm 0.38$ & $\mathbf{47.59 \pm 0.39}$ & $46.35 \pm 0.48$ & $46.41 \pm 0.18$ & $46.58 \pm 0.36$ \\
MINESWEEPER & $\mathbf{81.55 \pm 0.39}$ & $81.37 \pm 0.36$ & $81.42 \pm 0.32$ & $80.94 \pm 0.39$ & $80.63 \pm 0.24$ & $80.45 \pm 0.35$ \\
TOLOKERS & $\mathbf{79.26 \pm 0.61}$ & $79.22 \pm 0.68$ & TIMEOUT & TIMEOUT & $78.79 \pm 0.33$ & $79.13 \pm 0.54$ \\
\hline
\end{tabular}

\caption{Classification accuracies of GCN with AFR-3, AFR-4, BORF, SDRF, FoSR, or no rewiring strategy using our heuristics. Highest accuracies on any given dataset are highlighted in bold. 
}
\end{table}

\begin{table}[h!]
\centering
\tiny
\begin{tabular}{|l|c|c|c|c|c|c|}
\hline
 & 
\multicolumn{6}{|c|}{GIN} \\
\hline
DATA SET & AFR-3 & AFR-4 & BORF & SDRF & FOSR & NONE \\
\hline
AMAZON RATINGS & $48.02 \pm 0.29$ & $48.16 \pm 0.34$ & $\mathbf{48.70 \pm 0.26}$ & $47.81 \pm 0.45$ & $47.36 \pm 0.46$ & $47.66 \pm 0.36$ \\
MINESWEEPER & $79.12 \pm 0.31$ & $78.48 \pm 0.32$ & $\mathbf{79.81 \pm 0.29}$ & $78.69 \pm 1.65$ & $79.08 \pm 0.54$ & $78.19 \pm 0.36$ \\
TOLOKERS & $79.34 \pm 0.23$ & $\mathbf{79.51 \pm 0.39}$ & TIMEOUT & TIMEOUT & $78.81 \pm 0.39$ & $78.60 \pm 0.19$ \\
\hline
\end{tabular}

\caption{Classification accuracies of GIN with AFR-3, AFR-4, BORF, SDRF, FoSR, or no rewiring strategy using our heuristics. Highest accuracies on any given dataset are highlighted in bold. 
}
\end{table}

\newpage

\subsubsection{Gaussian mixtures for the ORC}
\label{subsubsection:orc_mixtures}

\begin{table}[h!]
\centering
\small
\begin{tabular}{|l|c|c|c|c|c|}
\hline
 & ENZYMES & IMDB & MUTAG & PROTEINS & PEPTIDES \\
\hline
$\Delta_L(\text{ORC})$ & $0.001 \pm 0.17$ & $0.032 \pm 0.011$ & $-0.371 \pm 0.127$ & $0.053 \pm 0.223$ & $-0.36 \pm 0.009$\\
\hline
\end{tabular}

\caption{Lower thresholds found for the graph classification datasets' ORC distributions using a Gaussian mixture model as described in section 4.}
\label{table:5}
\end{table}

\subsection{Additional figures}
\label{subsubsection:more_figures}

\subsubsection{Curvature Distributions}


\begin{figure*}[h]
    \centering
    {\includegraphics[width=.32\linewidth]{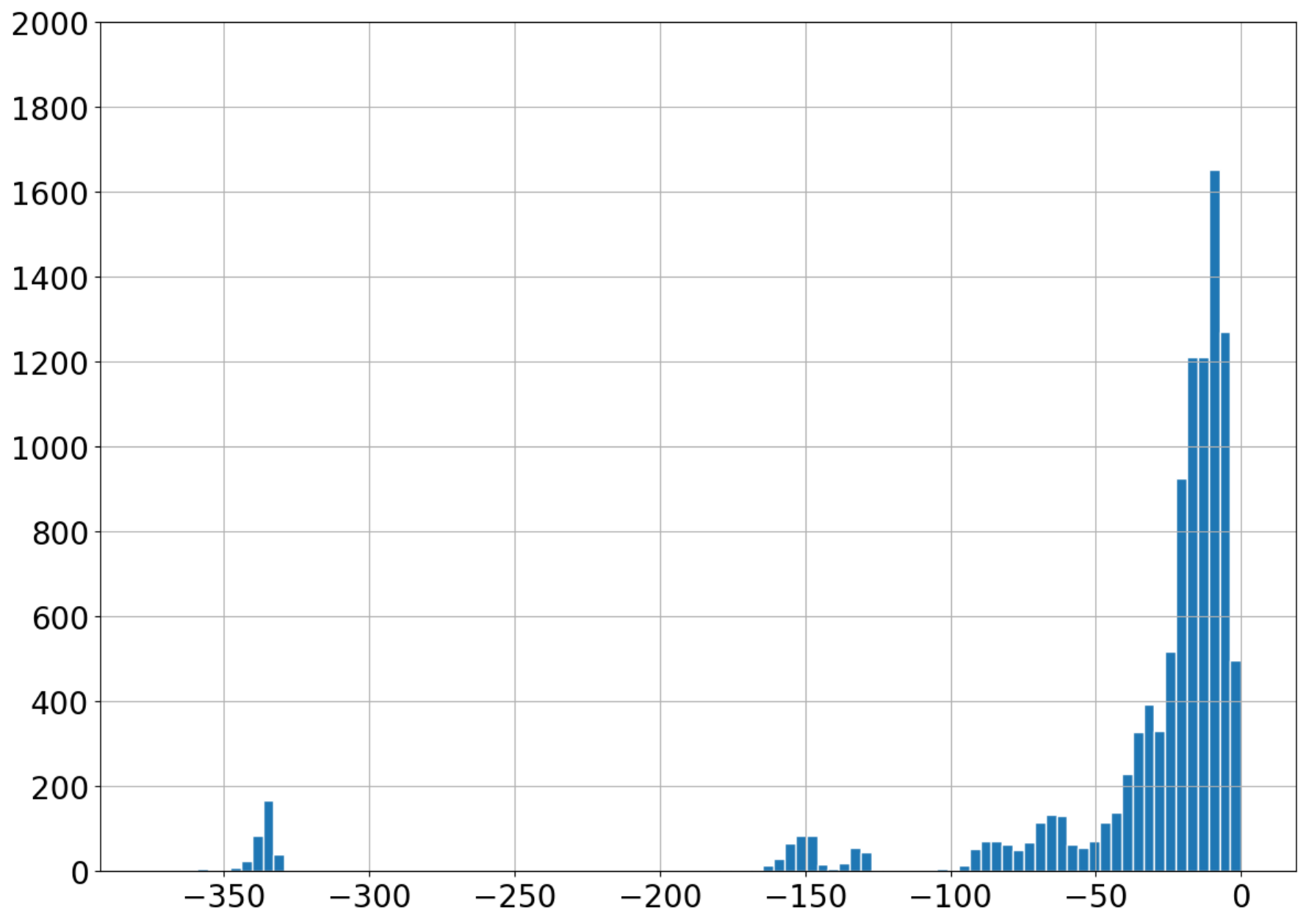}}
    \hfill
    {\includegraphics[width=.32\linewidth]{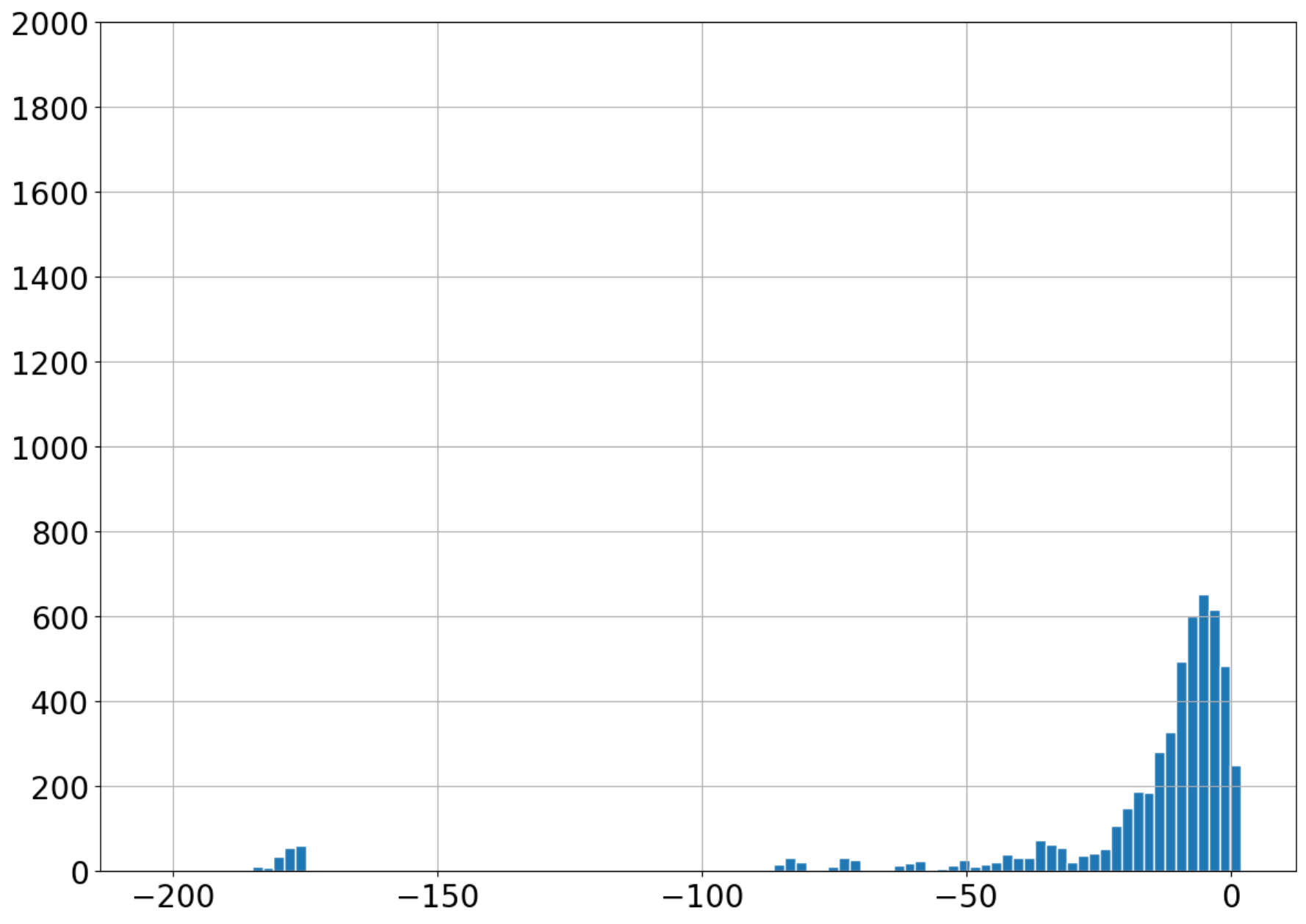}}
    \hfill
    {\includegraphics[width=.32\linewidth]{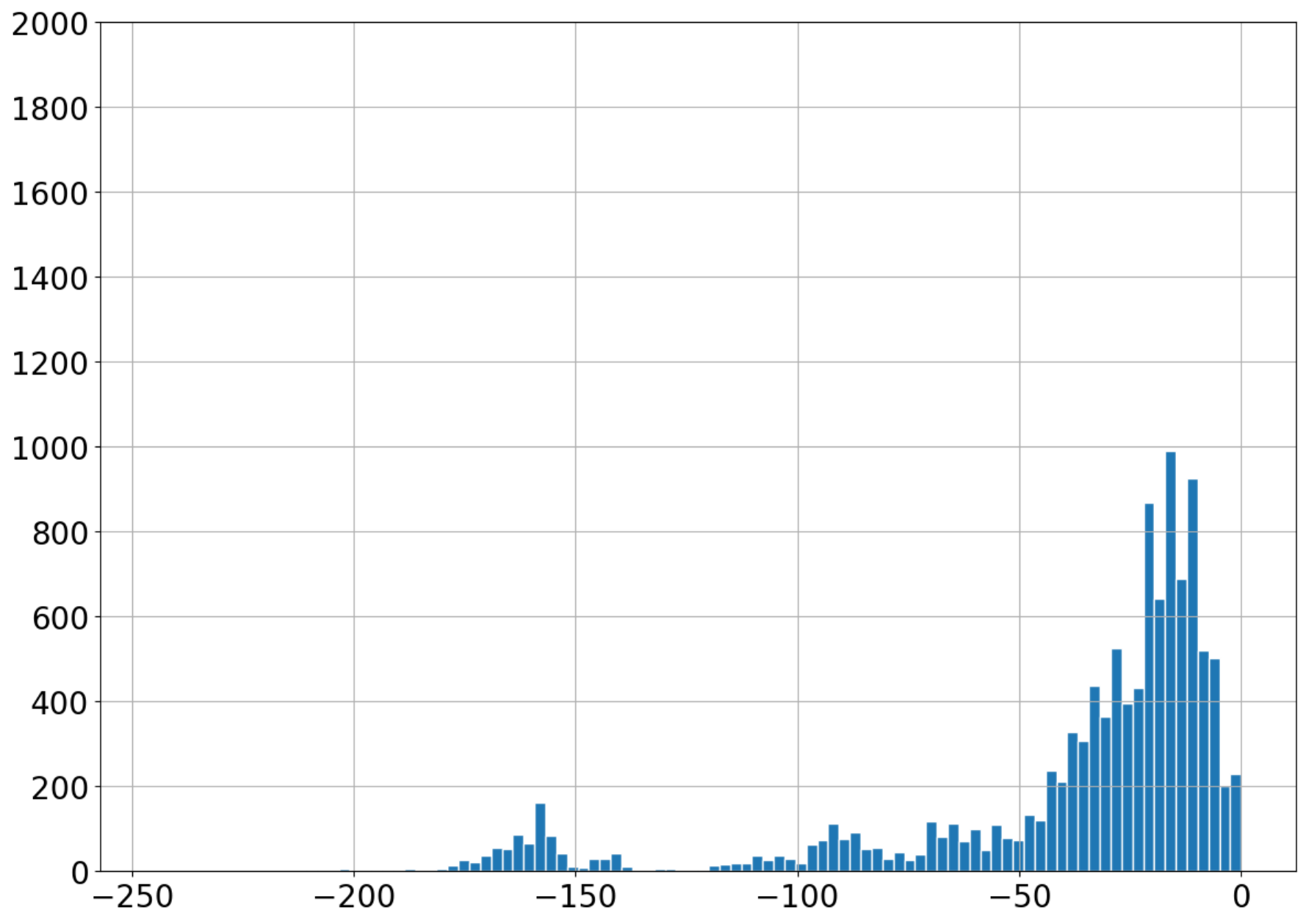}}
    \hfill
    \medskip
    {\includegraphics[width=.32\linewidth]{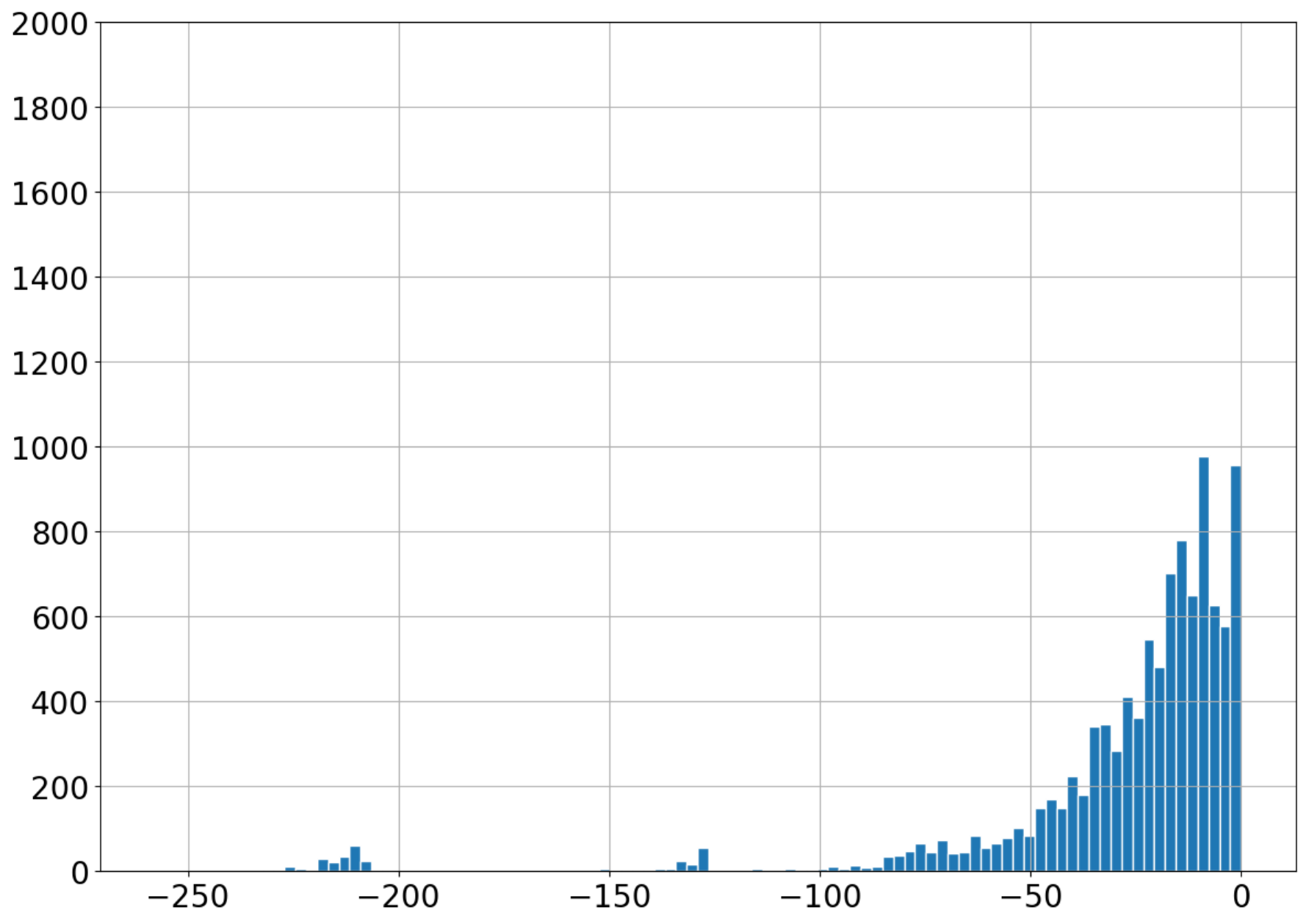}}
    \hfill
    {\includegraphics[width=.32\linewidth]{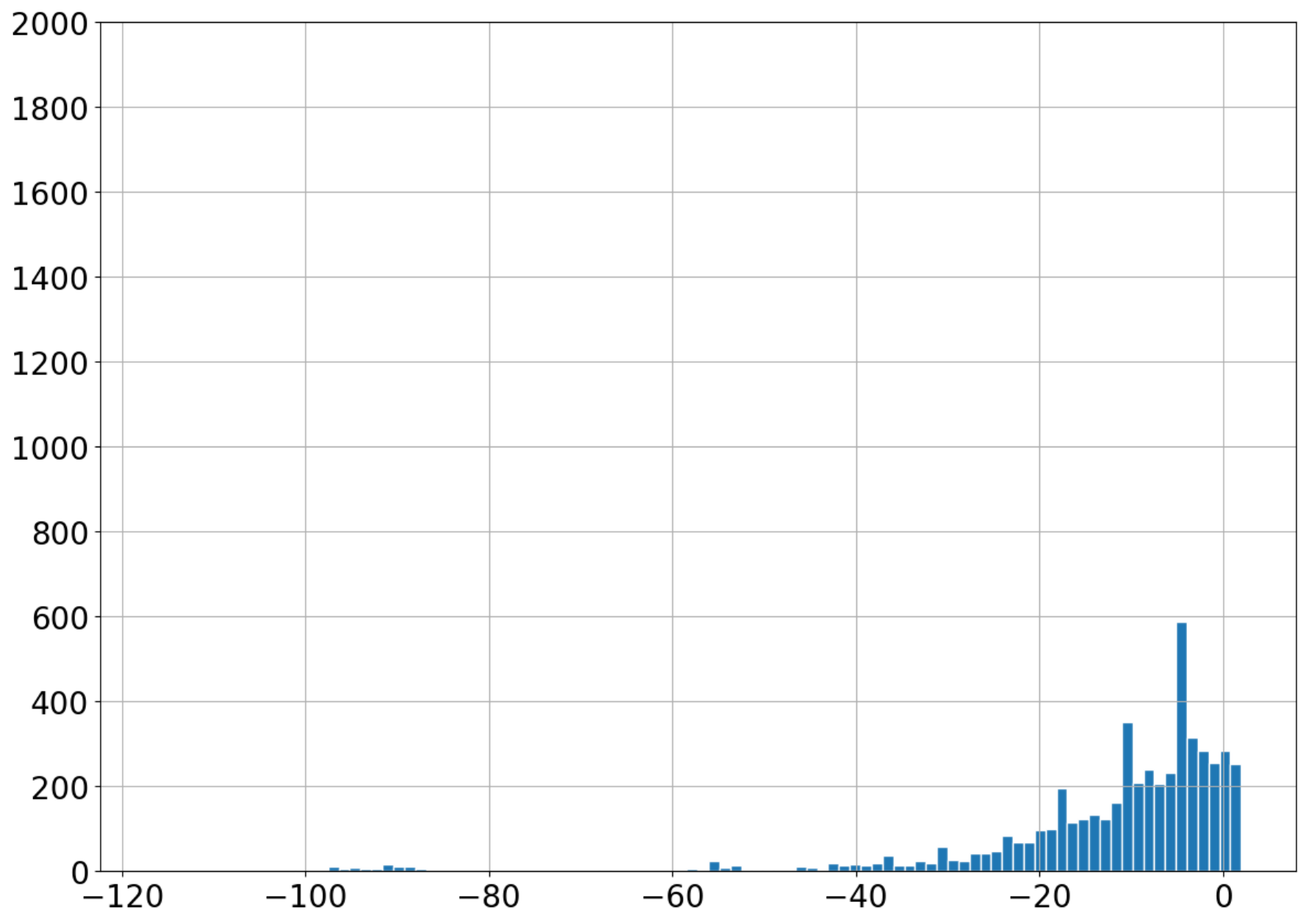}}
    \hfill
    {\includegraphics[width=.32\linewidth]{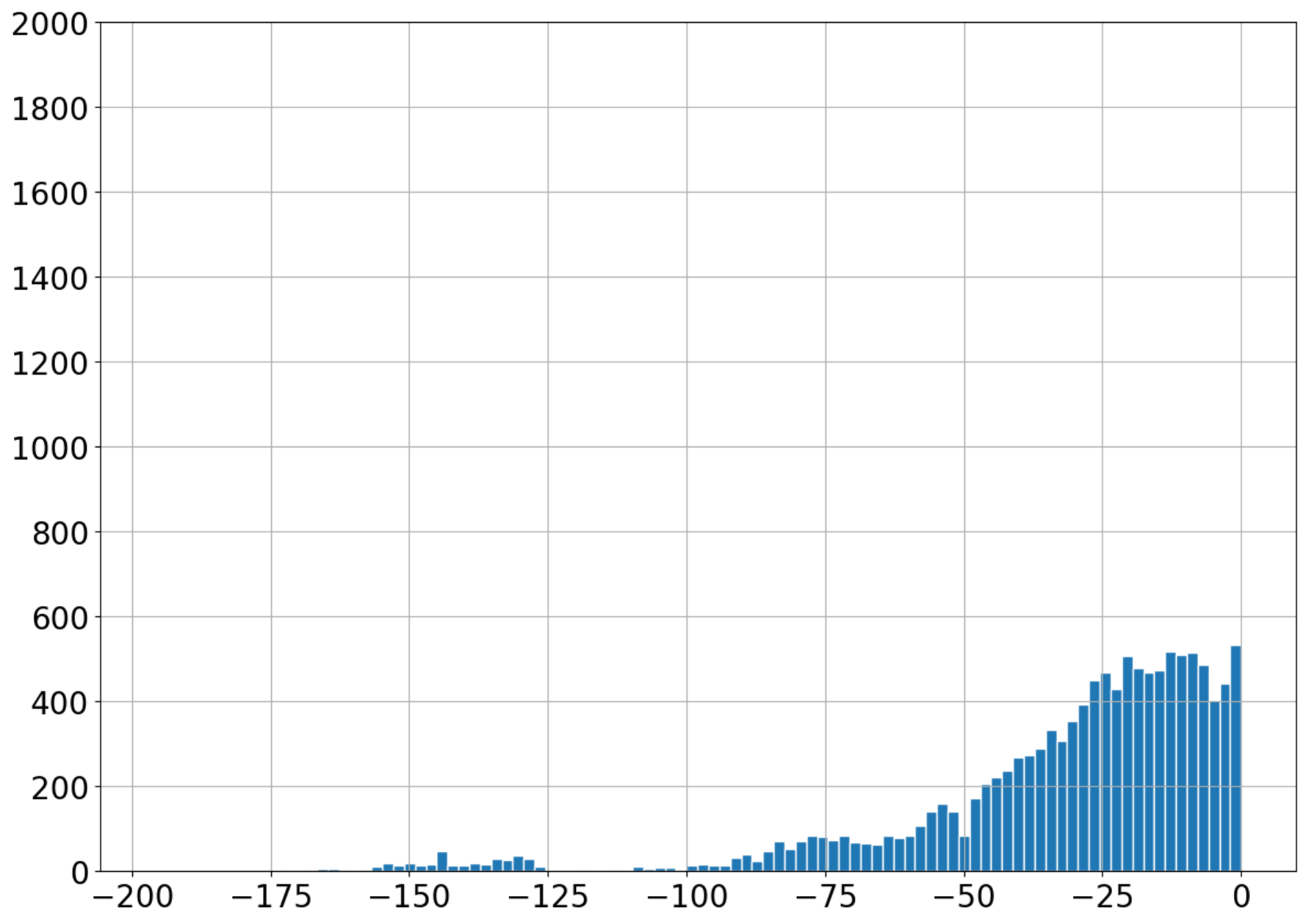}}
    \caption{AFRC distributions of edges in the Cora and Citeseer networks with no preprocessing (left), dropedge with $p = 0.5$ (center), and AFR-3 using our heuristics (right).}
\end{figure*}



\newpage

\subsubsection{Example Graphs}

\begin{figure*}[h]
    \centering
    {\includegraphics[width=.24\linewidth]{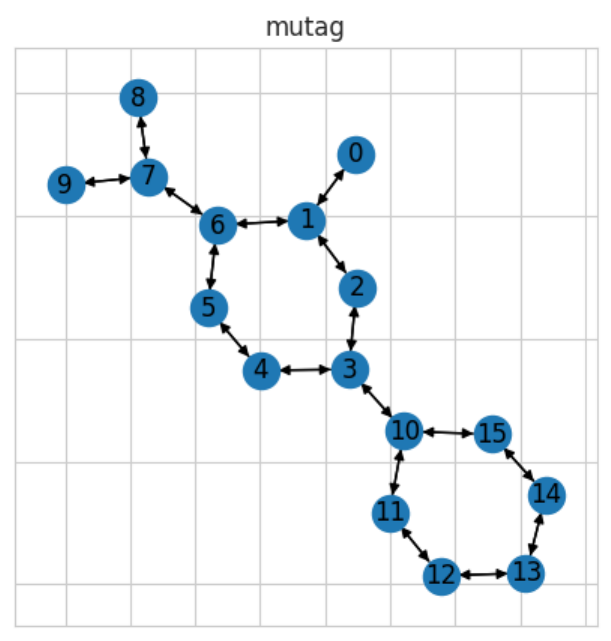}}
    \hfill
    {\includegraphics[width=.24\linewidth]{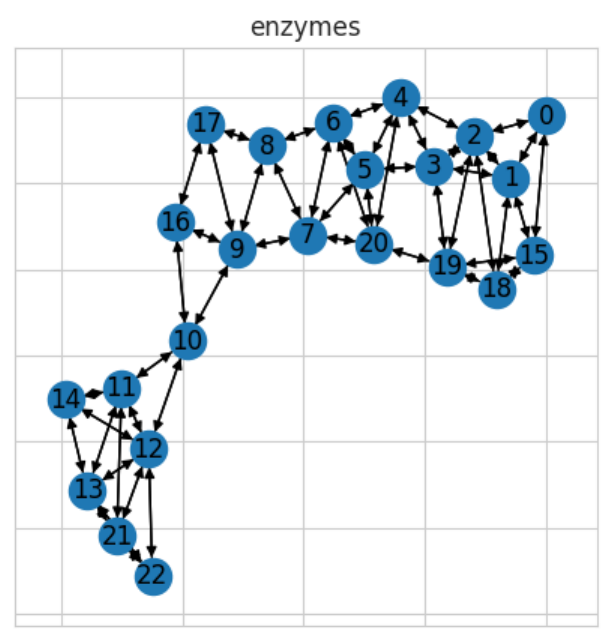}}
    \hfill
    {\includegraphics[width=.24\linewidth]{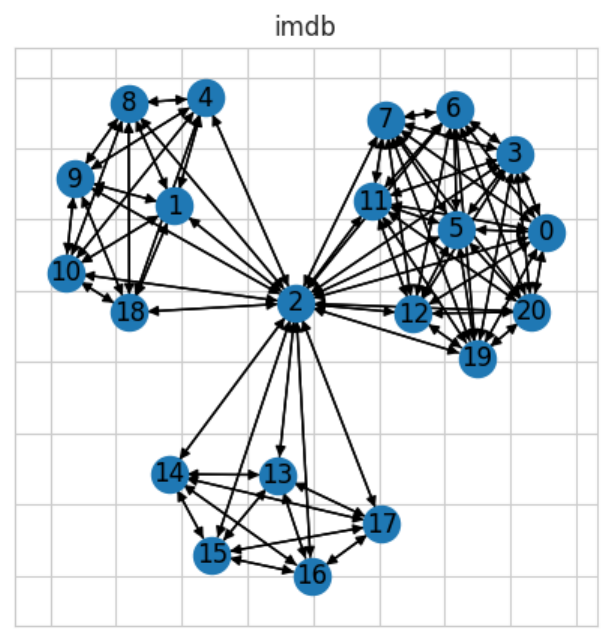}}
    \hfill
    {\includegraphics[width=.24\linewidth]{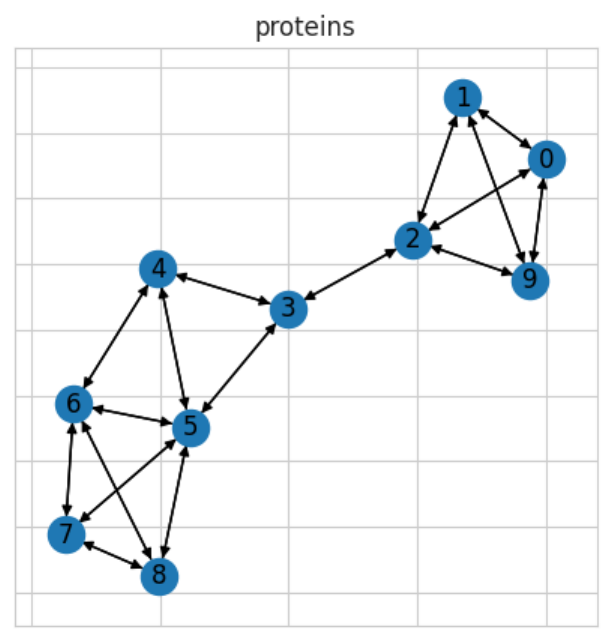}}\\
    \medskip
    {\includegraphics[width=.24\linewidth]{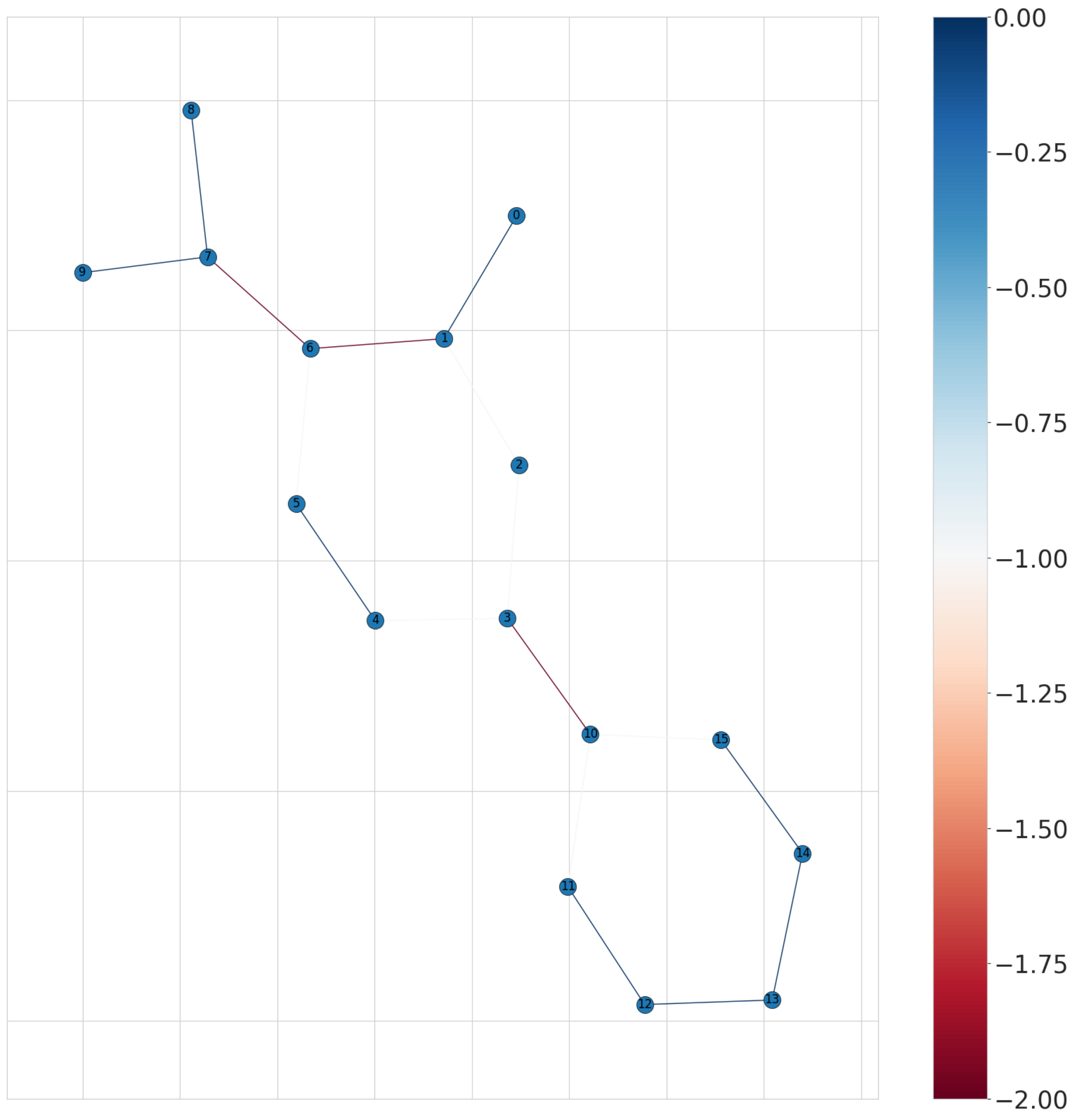}}
    \hfill
    {\includegraphics[width=.24\linewidth]{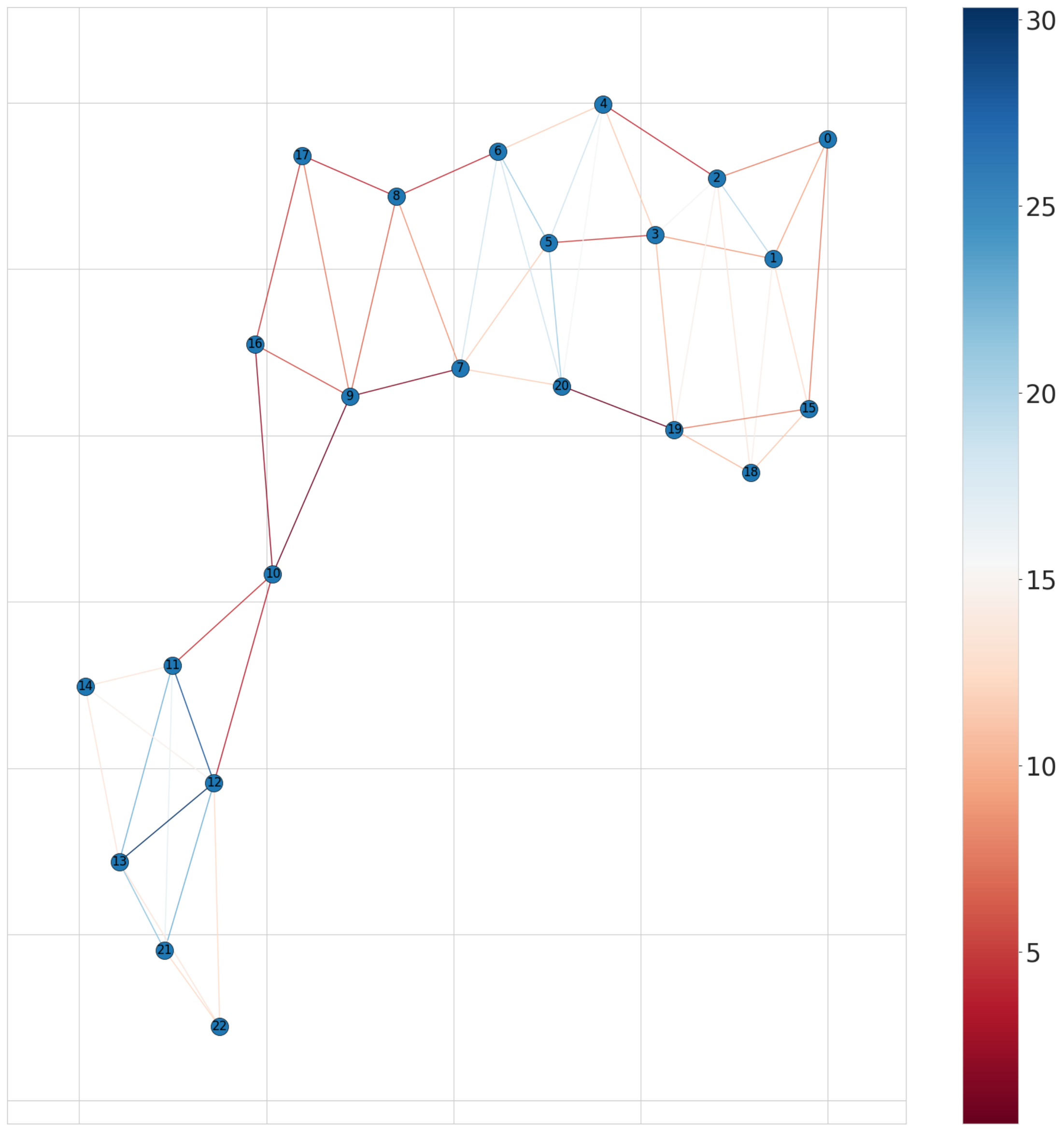}}
    \hfill
    {\includegraphics[width=.24\linewidth]{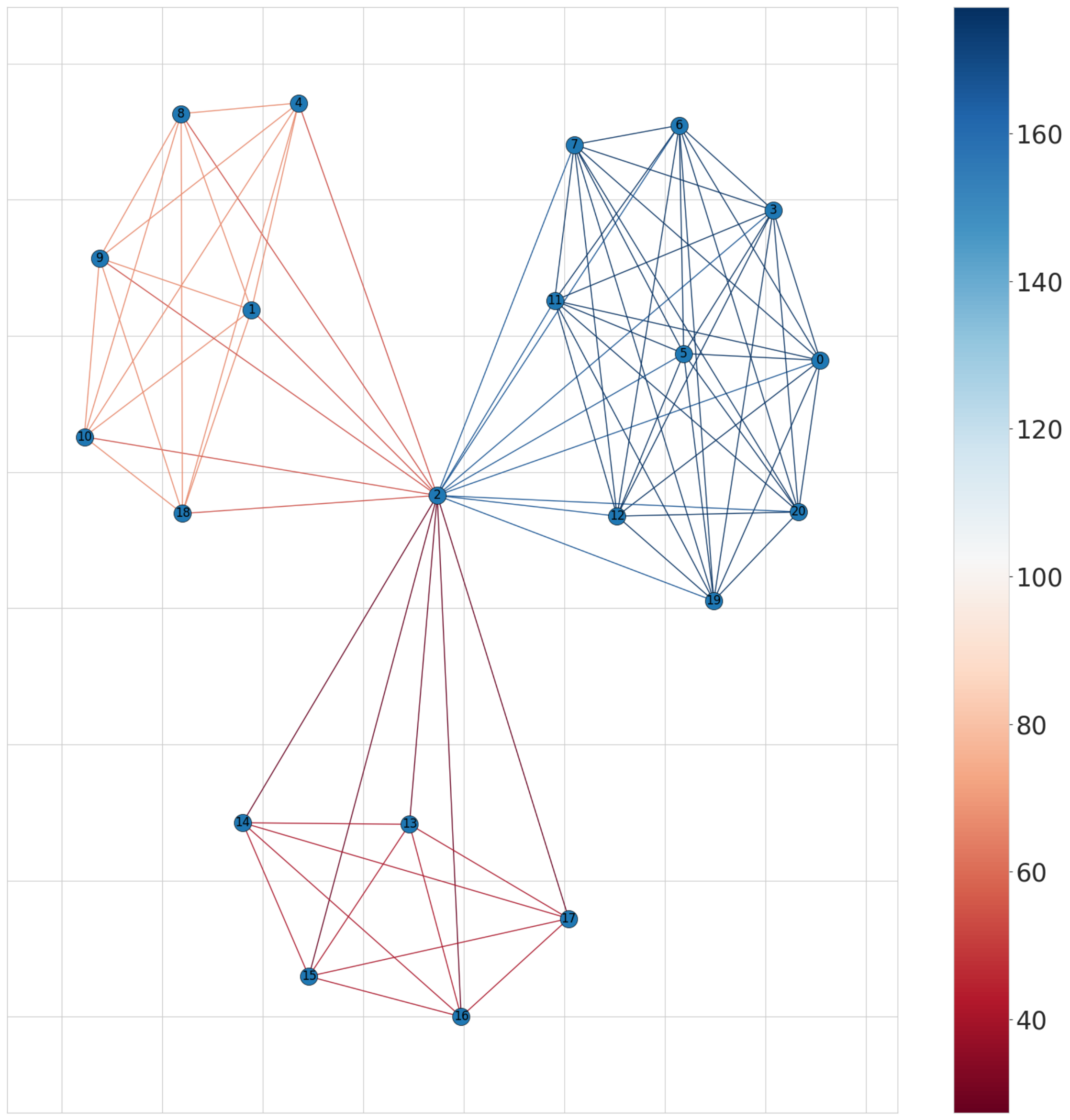}}
    \hfill
    {\includegraphics[width=.24\linewidth]{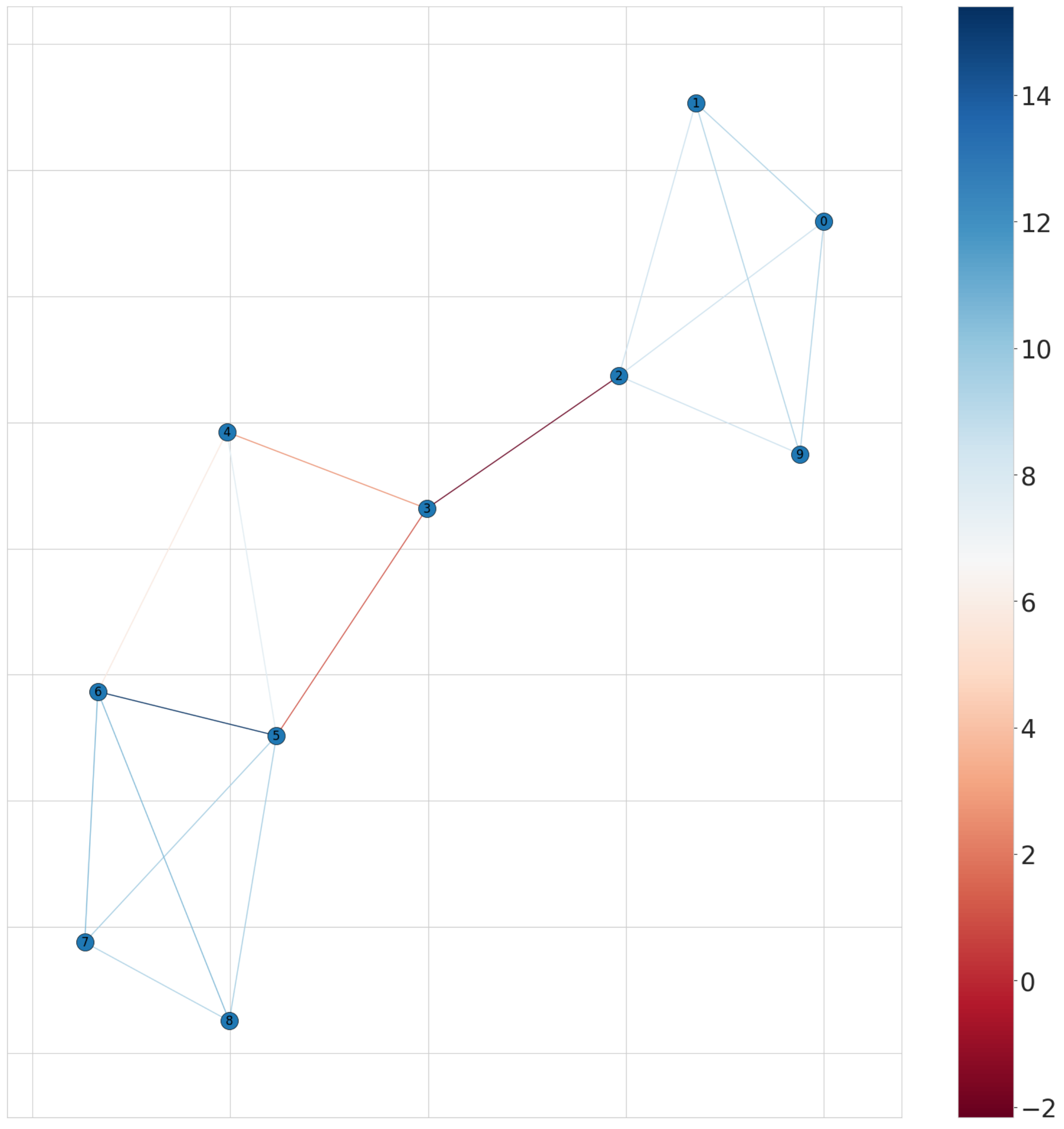}}
    \caption{Example networks from the mutag, enzymes, imdb, and proteins datasets, which we use for graph classification. The second row shows the same example networks with their edges colored according to their $\mathcal{AF}_3$ values. }\label{fig:afrc_examples}
\end{figure*}

\subsection{Statistics for datasets}
\label{subsection:datasets}

\subsubsection{General statistics for node classification datasets}

\begin{table}[h!]\label{tab:afrc_orc_heuristic}
\centering
\small
\begin{tabular}{|l|c|c|c|c|c|c|c|c|}
\hline
 & CORN. & TEX. & WISCON. & CORA & CITE. & CHAM. & COCO & PAS. \\
\hline
\# Graphs & 1 & 1 & 1 & 1 & 1 & 1 & 123286 & 11355 \\
\#NODES & 140 & 135 & 184 & 2485 & 2120 & 832 & 260-500 & 395-500 \\
\#EDGES & 219 & 251 & 362 & 5069 & 3679 & 12355 & 1404-2924& 2198-2890\\
\# FEATURES & 1703 & 1703 & 1703 & 1433 & 3703 & 2323 & 14* & 14* \\
\#CLASSES & 5 & 5 & 5 & 7 & 6 & 5 & 81 & 21 \\
DIRECTED & TRUE & TRUE & TRUE & FALSE & FALSE & TRUE & FALSE & FALSE \\
\hline
\end{tabular}

\caption{Statistics of node classification datasets. *See \cite{dwivedi2022LRGB} for details on LRGB datasets.}
\label{table:5}
\end{table}

\newpage

\subsubsection{Curvature distributions for node classification datasets}

\begin{table}[h!]\label{tab:afrc_orc_heuristic}
\centering
\small
\begin{tabular}{|l|c|c|c|c|c|c|c|c|c|}
\hline
 & 
\multicolumn{4}{|c|}{$\mathcal{AF}_3$} & \multicolumn{4}{|c|}{$\mathcal{AF}_4$} & \\
\hline 
DATASET & MIN. & MAX. & MEAN & STD & MIN. & MAX. & MEAN & STD & CORR.\\
\hline
CORA & -176 & 7 & -15.039 & 31.102 & -174 & 193 & -6.534 & 30.826 & 0.876 \\
CITESEER & -108 & 10 & -7.519 & 15.998 & -100 & 429 & 4.769 & 29.392 & 0.156 \\
TEXAS & -106 & 3 & -38.189 & 45.789 & -103 & 97 & -30.589 & 46.817 & 0.941 \\
CORNELL & -96 & 6 & -32.371 & 41.795 & -93 & 48 & -27.614 & 43.432 & 0.941 \\
WISCONSIN & -127 & 4 & -35.045 & 50.012 & -122 & 62 & -24.075 & 51.652 & 0.967 \\ 
\hline
COCO & $-8.1$ & $3$ & $-1.962$ & $1.424$ & $-1.7$ & $26.6$ & $7.853$ & $2.674$ & $0.407$\\
PASCAL & $-8.1$ & $2.9$ & $-1.984$ & $1.452$ & $-1.3$ & $24$ & $7.759$ & $2.502$ & $0.419$\\
\hline
\end{tabular}

\caption{Curvature statistics of node classification datasets.}
\label{table:6}
\end{table}

\subsubsection{General statistics for graph classification datasets}

\begin{table}[h!]\label{tab:afrc_orc_heuristic}
\centering
\small
\begin{tabular}{|l|c|c|c|c|c|}
\hline
 & ENZYMES & IMDB & MUTAG & PROTEINS & PEPTIDES \\
\hline
\#GRAPHS & 600 & 1000 & 188 & 1113 & 15535\\
\#NODES & 2-126 & 12-136 & 10-28 & 4-620 & 8-444\\
\#EDGES & 2-298 & 52-2498 & 20-66 & 10-2098 & 10-928\\
AVG \#NODES & 32.63 & 19.77 & 17.93 & 39.06 & 150.94\\
AVG \#EDGES & 124.27 & 193.062 & 39.58 & 145.63 & 307.30\\
\#CLASSES & 6 & 2 & 2 & 2 & 10 \\
DIRECTED & FALSE & FALSE & FALSE & FALSE & FALSE \\
\hline
\end{tabular}

\caption{Statistics of graph classification datasets.}
\label{table:5}
\end{table}

\subsubsection{Curvature distributions for graph classification datasets}

\begin{table}[h!]\label{tab:afrc_orc_heuristic}
\centering
\begin{tabular}{|l|c|c|c|c|c|c|c|c|c|}
\hline
 & 
\multicolumn{4}{|c|}{$\mathcal{AF}_3$} & \multicolumn{4}{|c|}{$\mathcal{AF}_4$} & \\
\hline 
DATASET & MIN. & MAX. & MEAN & STD & MIN. & MAX. & MEAN & STD & CORR.\\
\hline
MUTAG & -2.005 & 0.063 & -0.881 & 0.773 & -2.005 & 0.063 & -0.881 & 0.773 & 1 \\
ENZYMES & -5.152 & 4.570 & -0.273 & 2.230 & -3.182 & 20.258 & 6.448 & 4.923 & 0.687 \\
IMDB & -4.239 & 4.546 & 0.257 & 2.069 & -2.173 & 19.395 & 6.673 & 4.553 & 0.720 \\
PROTEINS & -4.777 & 8.039 & 2.917 & 3.562 & 66.097 & 191.539 & 120.982 & 25.424 & 0.661 \\ 
\hline
PEPTIDES & -1.999 & 0.994 & -0.671 & 0.773 & -.1996 & 0.995 & -0.671 & 0.773 & 0.999 \\
\hline
\end{tabular}

\caption{Curvature statistics of graph classification datasets.}
\label{table:6}
\end{table}

\noindent \textbf{Datasets.} For node classification, we conduct our experiments on the publicly available CORA, CITESEER \cite{DBLP:journals/corr/YangCS16}, TEXAS, CORNELL, WISCONSIN \cite{DBLP:journals/corr/abs-2002-05287} and CHAMELEON \cite{DBLP:journals/corr/abs-1909-13021} datasets. For graph classification, we use the ENZYMES, IMDB, MUTAG and PROTEINS datasets from the TUDataset collection \cite{DBLP:journals/corr/abs-2007-08663}. As a long-range task, we consider the PEPTIDES-FUNC
dataset from the LRGB collection \cite{dwivedi2022LRGB}.

\subsection{Hardware specifications and libraries}
\label{subsection:hardware}

\noindent All experiments in this paper were implemented in Python using PyTorch, Numpy PyTorch Geometric, and Python Optimal Transport. Figures in the main text were created using inkscape.

\noindent We conducted our experiments on a local server with the specifications laid out in the following table.

\begin{table}[h!]\label{tab:afrc_orc_heuristic}
\centering
\begin{tabular}{|l|l|}
\hline
COMPONENTS & SPECIFICATIONS \\
\hline
ARCHITECTURE & X86\_64 \\
OS & UBUNTU 20.04.5 LTS x86\_64 \\
CPU & AMD EPYC 7742 64-CORE \\
GPU & NVIDIA A100 TENSOR CORE \\
RAM & 40GB\\
\hline
\end{tabular}
\caption{}
\label{table:6}
\end{table}

\end{document}

%% file: macros.tex
\theoremstyle{definition}

\usepackage{booktabs} 
\usepackage{makecell} 

\numberwithin{equation}{section}